\documentclass[conference]{IEEEtran}
\usepackage{amsmath,amsfonts}
\usepackage{algorithmic}
\usepackage{array}
\usepackage[caption=false,font=normalsize,labelfont=sf,textfont=sf]{subfig}
\usepackage{textcomp}
\usepackage{stfloats}
\usepackage[hyphens]{url}
\usepackage{verbatim}
\usepackage{graphicx}
\usepackage{adjustbox}
\usepackage{multirow}
\usepackage{booktabs}
\usepackage{multicol}
\usepackage{threeparttable}
\usepackage{xcolor}
\def\BibTeX{{\rm B\kern-.05em{\sc i\kern-.025em b}\kern-.08em
    T\kern-.1667em\lower.7ex\hbox{E}\kern-.125emX}}
\usepackage{balance}

\usepackage{mathtools}
\usepackage[usestackEOL]{stackengine}
\usepackage{nicefrac}
\usepackage{amssymb}
\usepackage{pifont}
\newcommand{\cmark}{\ding{51}}%

\def\eg{\textit{e.g}. } 
\def\ie{\textit{i.e}. }

\def\wrt{w.r.t. } \def\etal{\textit{et al}. }

\begin{document}
\title{Privacy-preserving Multi-biometric Indexing based on Frequent Binary Patterns}

\author{\IEEEauthorblockN{Dail\'e Osorio-Roig, L\'azaro J. Gonz\'alez-Soler, Christian Rathgeb, Christoph Busch}
\IEEEauthorblockA{Biometrics and Security Research Group \\ Hochschule Darmstadt, Germany \\
\{daile.osorio-roig,lazaro-janier.gonzalez-soler,christian.rathgeb,christoph.busch\}@h-da.de}
}



\markboth{Journal of \LaTeX\ Class Files,~Vol.~18, No.~9, September~2020}%
{How to Use the IEEEtran \LaTeX \ Templates}

\maketitle

\begin{abstract}
The development of large-scale identification systems that ensure the privacy protection of enrolled subjects represents a major challenge. Biometric deployments that provide interoperability and usability by including efficient multi-biometric solutions are a recent requirement. In the context of privacy protection, several template protection schemes have been proposed in the past. However, these schemes seem inadequate for indexing (workload reduction) in biometric identification systems. More specifically, they have been used in identification systems that perform exhaustive searches, leading to a degradation of computational efficiency. To overcome these limitations, we propose an efficient privacy-preserving multi-biometric identification system that retrieves protected deep cancelable templates and is agnostic with respect to biometric characteristics and biometric template protection schemes. To this end, a multi-biometric binning scheme is designed to exploit the low intra-class variation properties contained in the frequent binary patterns extracted from different types of biometric characteristics. Experimental results reported on publicly available databases using state-of-the-art Deep Neural Network (DNN)-based embedding extractors show that the protected multi-biometric identification system can reduce the computational workload to approximately 57\% (indexing up to three types of biometric characteristics) and 53\% (indexing up to two types of biometric characteristics), while simultaneously improving the biometric performance of the baseline biometric system at the high-security thresholds. The source code of the proposed multi-biometric indexing approach together with the composed multi-biometric dataset, will be made available to the research community once the article is accepted.
\end{abstract}


\section{Introduction}
\IEEEPARstart{B}{iometric} technologies are rapidly gaining popularity due to their wide applicability. Biometric recognition of individuals based on distinctive biometric characteristics (BCs), \eg face or iris, is successfully deployed in many personal, commercial, and governmental identity management systems around the world, \eg border control, and national ID systems. A report on the global biometric market concerns the annual growth rate in biometric technologies by estimating 45.96 billion dollars in 2024~\cite{Pascu-GlobalBiometricMarket-2020}. In addition, biometrics vendors demand interoperability and deployment assuring maximum usability by including multimodal biometric solutions, \eg fight against fraud in banks~\cite{Beranek-MultiModalBanks-2023} and border and immigration~\cite{Burt-MultiModalBorderControl-2022} processes. These requirements (\ie interoperability and usability) motivate the development of biometric characteristic-agnostic systems. In particular, solutions that operate a common feature space while preserving high biometric performance by enabling new fusion schemes prior to any processing step in a biometric system. From an efficiency perspective, existing large-scale biometric systems are processing millions of subjects in the enrolment (\eg~\cite{UIDAI-AadhaarIssuedSummary-2018}) and re-enrolment processes (\eg~\cite{Nash-Re-EnrolProcess-2023}), respectively.

The above facts show the increase in computational cost. Also, the growth in monetary costs as large companies accelerate their large-scale processing by investing in advanced technologies (\eg hardware and speed-up devices). The challenging identification and duplicate enrolment check scenario where generally an exhaustive search (\ie one-to-many comparison) is a time-consuming task, demands practical solutions which are not dominated by the number of comparisons and hence a high computational workload. In recent years, significant interest has been raised in addressing this topic by investigating the \emph{workload reduction} (WR) methods~\cite{Drozdowski-WorkloadSurvey-IET-2019}, \eg~biometric indexing schemes, which have been introduced as methods with the aim of processing large amounts of biometric data with reasonable transaction times. 

In addition to the emerging topic of accelerating searches within large-scale biometric databases, the violation of data privacy came as a shock for many individuals (\eg period tracker scandal~\cite{Edwards-NewPeriodTrackerConcerns-2022}) as sensitive information (\eg personal health data) could be fully exposed. That is, in the context of a biometric system, unprotected storage of biometric references could lead to different privacy threats such as identity theft, linking across databases, or limited renewability~\cite{GomezBarrero-FrameworkForUnlinkability-TemplateProtection-2018}. Also,  privacy regulations, \eg the European Union (EU) General Data Protection Regulation 2016/679 (GDPR)~\cite{EU-Regulation-2016-679-on-DataPrivacy-160427}, usually define biometric information as sensitive data which requires strong mechanisms for the protection of stored data.

In the context of privacy protection, privacy-preserving biometric solutions have been challenged by natural intra-class variance of different biometric characteristics. Conventional cryptographic methods would require decryption of protected biometric data prior to the comparison step in order to prevent the effect of biometric variance in the encrypted domain. This is not the case with \emph{biometric template protection} schemes~\cite{Rathgeb-BTP-Survey-EURASIP-2011,Hahn-TemplateProtectionSurvey-2022} which enable a comparison of biometric data in the transformed domain (encrypted) and hence a permanent protection of biometric data. They are usually distinguished in the literature as \emph{cancelable biometrics} and \emph{biometric cryptosystems}. Generally, the latter category is not suggested in identification scenarios (where the workload is dominated by the typical exhaustive search-based), as they require complex comparison methods (\eg~\cite{Juels-FuzzyCommitment-1999,Juels-FuzzyVault-2006}), in contrast to cancelable schemes (\eg~\cite{Dong-face-identification-index-2020}).

Recently, Osorio-Roig \textit{et al.}~\cite{OsorioRoig-IndexingDeepCancelableTemplates-IJCB-2022} introduced the proof-of-concept of \emph{frequent binary patterns} for indexing deep cancelable face templates. This privacy-preserving solution allowed working on different cancelable protection schemes (\eg so-called BioHashing~\cite{Teoh-intro-biohashing-2005} and variants of Index-of-Maximum Hashing~\cite{Jin-index-of-hashing-2017}) ensuring a trade-off between computational workload and biometric performance for protected biometric identification systems. Motivated by our previous study (see \cite{OsorioRoig-IndexingDeepCancelableTemplates-IJCB-2022}), we present in this work (to the best of the authors’ knowledge) the \emph{first privacy-preserving multi-biometric identification system} based on the search of frequent binary patterns over cancelable biometric templates. The main contributions of the article are:

\begin{itemize}
    \item  An overview that delves into the area of computational workload reduction for the indexing of protected biometric templates in identification systems based on a single biometric characteristic.

    \item The successful application of the proof-of-concept of \emph{frequent binary patterns} on individual biometric characteristics, \ie face, iris, and fingerprint.

    \item An efficient privacy-preserving multi-biometric system that is agnostic across cancelable biometric template protection schemes (with binary representation) and biometric characteristics. This solution is able to operate on the most secure processing step (\ie feature level) in a biometric system by enabling fusion strategies on the concept of frequent binary patterns at two steps: the representation- and feature-based step. The fusion in the representation-step retrieval and indexing shows that the workload reduction and the biometric performance are irrespective of the ranking (\ie order of priority) of the biometric characteristics, in contrast to the fusion in the feature-step retrieval and indexing. 
    

    \item A thorough theoretical and empirical analysis of the trade-off between computational workload reduction and biometric performance of the proposed identification system on multi-modal large-scale datasets with state-of-the-art biometric recognition systems. Experimental evaluations compliant with the metrics defined in the ISO/IEC 19795-1:2021~\cite{ISO-IEC-19795-1-060401} show that a protected multi-biometric identification system can reduce the computational workload to approximately 57\% (indexing up to three types of biometric characteristics) and 53\% (indexing up to two types of biometric characteristics), while simultaneously improving the biometric performance at the high-security thresholds of a baseline biometric system. 
    
\end{itemize}

The remainder of this work is organised as follows: related works summarising concepts related to information fusion, workload reduction and biometric template protection are revisited in Sect.~\ref{sec:works}. In Sect.~\ref{sec:system}, the proposed system is described in detail. Sect.~\ref{sec:experimental-setup} presents the experimental evaluations and results are reported and discussed in Sect~\ref{sec:results}. Finally, conclusions are drawn in Sect.~\ref{sec:conclusions}. 

\section{Related works}
\label{sec:works}

This section describes the background and related work on reducing computational workload in protected biometric identification systems. Whereas Sect.~\ref{sec:fusion-works} introduces the fusion strategies commonly used in biometrics, Sect.~\ref{sec:workload-works} addresses the problem of workload reduction on biometric systems. Finally, key work related to the \textit{workload-reduction} and \textit{biometric identification systems} areas on biometric template protection is summarised in Sect.~\ref{sec:protection-works}. 

\subsection{Biometric Information fusion}
\label{sec:fusion-works}

Biometric information fusion allows combining biometric data at different levels of processing in a biometric system. Those systems which enable biometric information fusion are known in the literature as multi-biometric systems. Generally, multi-biometric schemes combine or fuse multiple sources of information to improve the overall discriminative power of a single biometric recognition system~\cite{Drozdowski-CascadingFiltering-TBIOM-2020}. The fusion strategies can be categorised in the biometric context as multi-types, multi-sensorial, multi-algorithms, multi-instances, and multi-presentations~\cite{RossNandakumarJain-HandbookOfMultibiometrics-2006,ISO-IEC-24722-TR-Fusion-150216}.  






The system proposed in this work relates to the first scenario, \ie multi-type, which relies on the fusion of different types of BCs (\eg facial and iris images). Specifically, three types of BCs are selected and subsequently utilised in a binning and fusion scheme. Note that given the simplicity of the proposed scheme, other fusion categories, such as multi-sensorial, multi-instances and multi-presentations can be also employed. In addition to the general categories above, several levels of biometric processing can be distinguished at which information fusion can take place~\cite{RossNandakumarJain-HandbookOfMultibiometrics-2006,ISO-IEC-24722-TR-Fusion-150216}: sensor, feature, score, rank, and decision.

In the scope of this article, the fusion of information from multiple features and at the score level is of major interest, as the proposed scheme in Sect.~\ref{sec:system} is designed to operate at those levels of the biometric processing pipeline. The feature-level fusion has been also considered, as it is among the most convenient techniques contributing to the highest privacy protection and security level, respectively~\cite{Merkle-Multi-modal-instanceFusion-BIOSIG-2012,Rathgeb-BTP-Survey-EURASIP-2011}. 

Information fusion in biometrics has been widely addressed in the scientific literature. An interested reader is therefore referred to, \eg   Ross \textit{et al.}~\cite{RossNandakumarJain-HandbookOfMultibiometrics-2006} for a general introduction to this topic and  Paul~\textit{et al.}~\cite{Paul-Decision-Multi-modal-2014} for score-level fusion specifically, as well as Dinca and Hancke~\cite{Dinca-Fusion-Survey-2017}, Singh~\textit{et al.}~\cite{Singh-ComprehensiveSurvey-2019}, and ISO/IEC TR 24722~\cite{ISO-IEC-24722-TR-Fusion-150216} for more recent works relating to the general topic of biometric information fusion.

\subsection{Computational workload reduction}
\label{sec:workload-works}

Biometric identification systems require fast response times, as the typical exhaustive search-based retrieval method demands high computational costs. Thus, the computational complexity tends to grow linearly with the number of enrolled data subjects~\cite{Daugman-BiometricDesignLandscapes-2000}. As expected, the investment in expensive hardware that contributes to the parallel processing/distribution can be used to maintain a quick response time in a biometric identification transaction. Whereas many companies spend high monetary costs to achieve the desired times, one possibility that is often overlooked is the optimisation of the underlying software and/or algorithms. In this context, a solution to said problems (\ie high computational and monetary costs) is the research field of \textit{computational workload reduction} which allows decreasing the dependence on the investment of the physical infrastructure and focusing more attention on the software and/or algorithms. Workload reduction-based methods work directly on the optimisation of the amount of computations required for some specific tasks in the biometric processing pipeline. For instance, for a biometric identification transaction, the computational costs at the biometric template comparison level typically dominate the computational effort of the entire system. Thus, most of these methods have been categorised in~\cite{Drozdowski-WorkloadSurvey-IET-2019} as \textit{pre-selection} approaches. These methods seek to reduce the number of biometric template comparisons (\ie reducing the search space (see \eg~\cite{Kavati-SurveySearchReduction-2018})), and \textit{feature transformation}, aimed at accelerating the computational cost produced in a one-to-one comparison (see \eg~\cite{Drozdowski-DeepFaceBinarisation-ICIP-2018}). The former is of interest in the context of this article. For further information on such methods, the reader can be referred to~\cite{Drozdowski-WorkloadSurvey-IET-2019}. 

Naturally, those workload reduction-based techniques (\ie pre-selection methods) have achieved decreasing the search spaces \wrt the typical exhaustive searches. Conceptually, such approaches are mostly custom-built for specific biometric systems, \eg single biometric characteristics or feature extractors introducing specific representations, and are not expected to be applicable within other systems, \eg containing different types of biometric characteristics to be processed. In addition, they are primarily designed to facilitate the reduction of the computational workload associated with biometric identification transactions in unprotected biometric systems (\ie unprotected template indexing), which are prone to unauthorised attacks. The latter has motivated the scientific literature to investigate new customised procedures capable of performing the protected template indexing while reducing the overall computational effort per biometric identification transaction.

\begin{table*}[!t]
	\scriptsize
	\begin{center}
	\caption{Most relevant approaches on biometric template protection for biometric identification systems based on a single biometric characteristic. Results reported for best configurations and scenarios}	\label{tab:related-works}
     \begin{adjustbox}{max width=\textwidth}
			  \begin{threeparttable}

	 \begin{tabular}{c c c c c c c c} \toprule
		 \textbf{Approach} & \textbf{\shortstack{WR \\ category}} & \textbf{\shortstack{BTP \\ category}} & \textbf{\shortstack{Biometric \\ characteristics}} & \textbf{\shortstack{Biometric \\ performance}} & \textbf{\shortstack{Efficient \\ comparison}} \\ \midrule
		
		 \multirow{1}{*}{Wang \textit{et al. }~\cite{Wang-indeference-similarity-2017}} & \Centerstack{Pre-selection,  \\ Feature transformation } & \multirow{1}{*}{Non-traditional BTP} & Face & 95\% H-R 	&\multirow{1}{*}{ (\cmark)}&  \\ \midrule

		Murakami \textit{et al. }~\cite{Murakami-cancelable-identification-2019} & Feature transformation & Cancelable biometrics & Face & 0.1\% FRR, 0.022\% FAR& (\cmark)\\ \midrule

		\multirow{1}{*}{Dong \textit{et al. }~\cite{Dong-face-identification-index-2020}} & \multirow{1}{*}{Feature transformation}	&
		\multirow{1}{*}{Cancelable biometrics} & Face &  99.75\% R-1 &\multirow{1}{*}{(\cmark)}\\	\midrule

  	\multirow{1}{*}{Osorio-Roig \textit{et al.}~\cite{OsorioRoig-IndexingDeepCancelableTemplates-IJCB-2022}}&\multirow{1}{*}{Pre-selection} &\multirow{1}{*}{Cancelable biometrics}&Face &$\sim$99.00\% H-R & \multirow{1}{*}{\cmark} \\ \midrule

    Drozdowski \textit{et al.}~\cite{Drozdowski-BloomFilterIndexing-IET-2017}&Pre-selection&Cancelable biometrics&Iris&0.1\% FPIR,93.21–97.50\% FNIR&\cmark \\ \midrule

     Choudhary \textit{et al.}~\cite{Choudhary-ProtectionFingerVein-2022}&Feature transformation&Cancelable biometrics&Finger-Vein&99.70\%R-1&(\cmark)\\ \midrule
	
	\multirow{1}{*}{Sardar \textit{et al. }~\cite{Sardar-novel-cancelable-face-hashing-2020}} & \multirow{1}{*}{Feature transformation}	& \multirow{1}{*}{Cancelable biometrics} & Face & 99.85\% CRR-1 &\multirow{1}{*}{(\cmark)}\\ \midrule
															   
Drozdowski \etal~\cite{Drozdowski-HomomorphicIdentificationFace-BIOSIG-2019} 	& Feature transformation & Homomorphic encryption	& Face & $\sim$5\% FNIR, 1\% FPIR &(\cmark) \\					   \midrule

	Engelsma \textit{et al. }~\cite{Engelsma-homomorphic-representation-search-2020} & Feature transformation & Homomorphic encryption	& Face & 81.4\% R-1 &(\cmark) \\ \midrule
	
	\multirow{1}{*}{Osorio-Roig \textit{et al.}~\cite{OsorioRoig-StableHashFaceIdentification-TBIOM-2021}} & \multirow{1}{*}{Pre-selection} & \multirow{1}{*}{Homomorphic encryption}	&Face  & 1.0\% FPIR, 2.5\% FNIR & \multirow{1}{*}{\cmark}&\\		\midrule																																								

Drozdowski \textit{et al.}~\cite{Drozdowski-FeatureFusionIndexing-ACCESS-2021} & Pre-selection & Homomorphic encryption&Face& 0.1\% FPIR, 0.42\% FNIR & \cmark	\\ \midrule

Kolberg \textit{et al.}~\cite{Kolberg-IrisNTRU-BTP-WIFS-2019}&Feature transformation& Homomorphic encryption& Iris&98.08\% R-1&(\cmark) \\ \midrule

Bauspiess \textit{et al.}~\cite{Bauspiess-KeywordSearch-2022}&Pre-selection&Homomorphic encryption&Face&0.1\% FPIR,~1.2\% FNIR&\cmark \\ \midrule

 Engelsma \textit{et al.}~\cite{Engelsma-FixedLengthRepresentation-2019}&\Centerstack{Pre-selection,\\ Feature transformation }&Homomorphic encryption&Fingerprint&99.93\% H-R&\cmark\\ \midrule

\multirow{1}{*}{Dong \textit{et al.}~\cite{Dong-secure-fuzzy-vault-2021}}&\multirow{1}{*}{ Feature transformation }&\multirow{1}{*}{Fuzzy vault}& Face &99.86\% R-1& \multirow{1}{*}{(\cmark)} \\

 \bottomrule																																														   																												  
																  
\end{tabular} 	
\begin{tablenotes}
  \item H-R: Hit Rate, FRR: False Rejection Rate, FAR: False Acceptance Rate, R-1: Rank-1 Identification Rate, DIR: Detection and Identification Rate, CRR: Correct Recognition Rate at Rank-1, FPIR:False Positive Identification Rates, FNIR: False Negative Identification Rates, \cmark: Property fulfilled, (\cmark): Property partially fulfilled.
\end{tablenotes}
\end{threeparttable} 
\end{adjustbox}
    \end{center}
\end{table*}

\subsection{Biometric template protection}
\label{sec:protection-works}
\textit{Biometric template protection} schemes allow protecting biometric references (\ie biometric templates) in an unprotected storage environment of a biometric system. Once they are protected, a set of properties are expected to be inherent to the transformed or protected templates constraining the flexibility of the biometric processing pipeline compared to unprotected templates. Comprehensive surveys on this field can be found in~\cite{Rathgeb-BTP-Survey-EURASIP-2011,ISO-IEC-24745-TemplateProtection-2022,Hahn-TemplateProtectionSurvey-2022}. Generally, template protection methods are categorised as \textit{cancelable biometrics} and \textit{biometric cryptosystems}. The former employ transformations in the signal or feature domain that allow biometric comparison in the transformed (encrypted) domain~\cite{Patel-CancelableReview-2015}. The latter (\eg fuzzy vault schemes~\cite{Juels-FuzzyVault-2006}) usually bind a key to a biometric feature vector resulting in a protected template. Thus, the biometric comparison is then performed indirectly by verifying the correctness of a retrieved key~\cite{Uludag-CryptosystemsChallenges-2004}. In particular, homomorphic encryption-based template protection schemes are distinguished as biometric cryptosystems whose specific designs allow computing operations directly in the encrypted domain with results comparable to those in the plaintext domain (\ie unprotected domain)~\cite{Aguilar-Melchor-HomomorphicEncryptionAdvances-2013}. The challenge of unprotected templates being replaced by protected templates leads to requirements or properties which must be fulfilled according to ISO/IEC IS 24745~\cite{ISO-IEC-24745-TemplateProtection-2022}:
\begin{LaTeXdescription}
\item \textit{Irreversibility}: The infeasibility of reconstructing the original biometric sample given a protected template. This type of property guarantees the privacy of the users’ data (\eg avoiding dislocating the subject's ethnic information) and additionally, the security of the system is increased against \eg presentation attacks and face reconstruction from deep templates. 
\item \textit{Unlinkability}: The infeasibility of determining if two or more protected templates were derived from the same biometric instance, e.g. face. By fulfilling this property, cross-matching across different databases is prevented.
\item \textit{Renewability}: The possibility of revoking old protected templates and creating new ones from the same biometric instance and/or sample, \eg face image. With this property fulfilled, it is possible to revoke and re-generate new templates in case the database is compromised. 
\item \textit{Performance preservation} the requirement of the biometric performance not being significantly impaired by the protection scheme.
\end{LaTeXdescription}

Tab.~\ref{tab:related-works} lists the most relevant scientific works on biometric template protection for biometric identification systems based on a single biometric characteristic. The approaches have been analysed in terms of efficient comparison (\ie workload reduction) and biometric performance. Scientific works on biometric cryptosystems for identification~\cite{Engelsma-homomorphic-representation-search-2020,Drozdowski-HomomorphicIdentificationFace-BIOSIG-2019,Kolberg-IrisNTRU-BTP-WIFS-2019} have been commonly focused on providing evidence of practical applicability. The majority of them have contributed to reducing the effort at a one-to-one comparison level by feature transformation while other approaches~\cite{OsorioRoig-StableHashFaceIdentification-TBIOM-2021,Bauspiess-KeywordSearch-2022} worked on the reduction of one-to-many comparisons. It is well-known that cancelable schemes appeared to be more suitable in an identification scenario~\cite{OsorioRoig-IndexingDeepCancelableTemplates-IJCB-2022}, in contrast to biometric cryptosystems (\eg~\cite{Dong-secure-fuzzy-vault-2021}). That is due to the fact that the design of cancelable biometrics does not require comparison strategies that usually enable the non-flexibility of launching non-arithmetic operations~\cite{Sperling-HEFusionMultibiometric-2022} or verifying the correctness of a retrieved key~\cite{Uludag-CryptosystemsChallenges-2004}. From a practical perspective, cancelable approaches have been therefore successfully considered over identification scenarios for different biometric characteristics (\eg face, iris, and fingerprint). As mentioned above, these schemes introduce non-invertible transformations at the feature level which usually allow retaining efficient biometric comparators of the corresponding unprotected systems. This way, the majority of published cancelable schemes applied transformations in the feature domain while maintaining acceptable biometric performance and low computational workload. Over the past years, some feature transformations (\eg BioHashing~\cite{Sardar-novel-cancelable-face-hashing-2020}) covered discriminative power-based gaps addressing the indexing protected templates with an identification rate at the rank 1 (R-1). Also, the locality sensitive hashing (LSH)~\cite{Indyk-dimensionality-curse-1998} nature has recently been exploited and designed to obtain compact non-invertible features (\eg~\cite{Dong-face-identification-index-2020,Choudhary-ProtectionFingerVein-2022}) where similarly protected templates are more likely to have the same hash collision compared to dissimilar ones. 

The described solutions applied workload reduction through an acceleration of a one-to-one comparison. In contrast, other researchers (\eg~\cite{Drozdowski-BloomFilterIndexing-IET-2017,OsorioRoig-IndexingDeepCancelableTemplates-IJCB-2022}) have explored computational workload reduction to decrease the number of one-to-many comparisons which dominates the overall computational effort in biometric identification transactions~\cite{Drozdowski-FeatureFusionIndexing-ACCESS-2021}. More precisely, Osorio-Roig \textit{et al}~\cite{OsorioRoig-IndexingDeepCancelableTemplates-IJCB-2022} proposed recently the retrieval of cancelable deep face templates based on their frequent binary patterns. The design of this type of retrieval enabled the use of different cancelable biometric template protection schemes. To sum up, all published works on cancelable biometric template protection for biometric identification worked on an exhaustive search when only feature transformation was employed. Whereas other works reduced the one-to-many search (\ie pre-selection-based approaches), such schemes are usually not flexible or not designed to work on different biometric characteristics. In addition, some generic multi-biometric indexing methods suitable to work only on unprotected domains have been proposed \eg in~\cite{Drozdowski-CascadingFiltering-TBIOM-2020,Jayaraman-MultimodalIndexing-2008,Gyaourova-CodingSchemeMultimodal-2009}.

\section{Proposed system}
\label{sec:system}

\begin{figure*}[!t]
    \centering
    \includegraphics[width=0.9\linewidth]{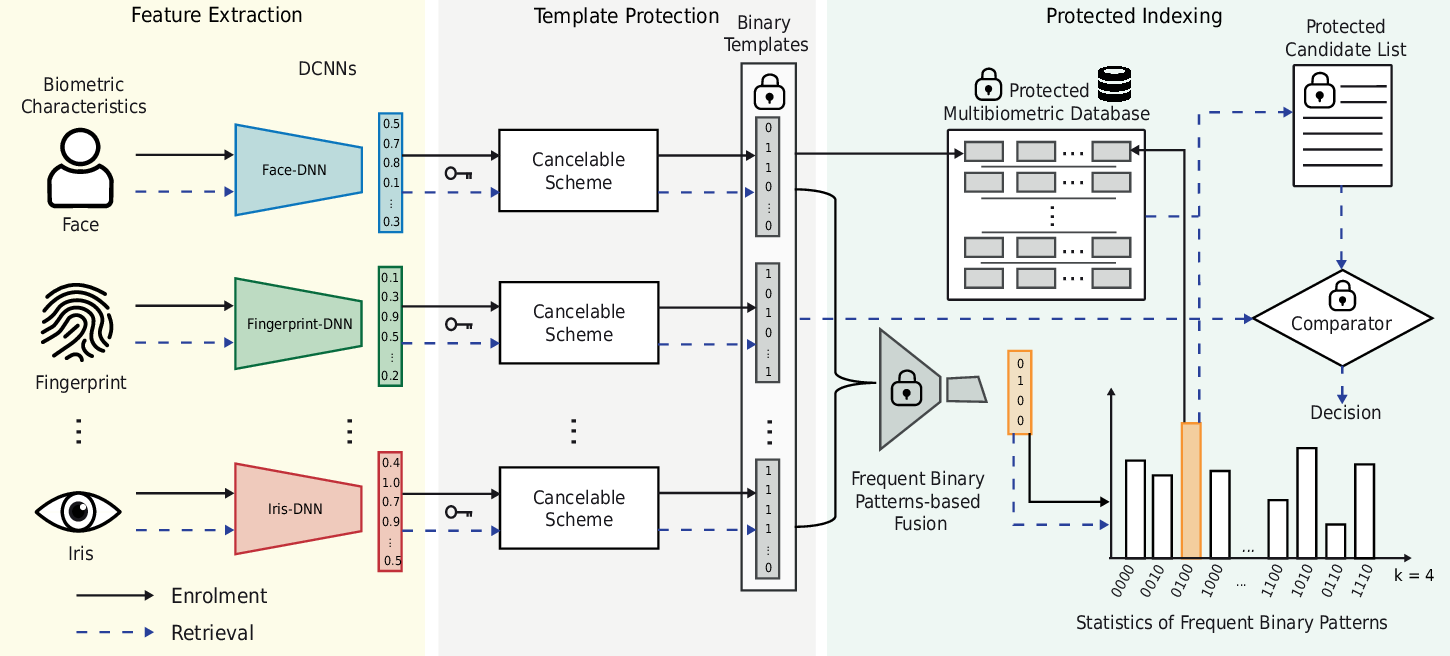}
    \caption{Conceptual overview of the proposed multi-biometric scheme. Firstly, the system receives different biometric characteristics or instances which are processed by state-of-the-art DNN-based embedding extractors. Subsequently, feature vectors of equal size are protected and encoded in a binary representation by well-established cancelable schemes. Techniques of information fusion are then applied to the protected features along with the concept of frequent binary patterns for the indexing and retrieval steps. Finally, a protected candidate list can be returned taking into account the statistics of their frequent patterns.}   
    \label{fig:overview}
\end{figure*}

Consider a biometric enrolment database containing references protected by cancelable schemes\footnote{We assume these schemes yield features containing binary representations.} of $N$ data subjects for $m$ different biometric characteristics or instances. A trivial search process for a single biometric identification transaction would be to conduct the comparisons exhaustively, \ie the workload of a baseline system is estimated as $W_{baseline} = N\cdot m $ comparisons for all biometric characteristics. In fact, for an improvement of the biometric performance or workload reduction (see~\cite{Drozdowski-CascadingFiltering-TBIOM-2020}), \eg fusing the scores using one of the traditional strategies (such as score or rank level fusion) mentioned in Sect.~\ref{sec:fusion-works}, the workload would be dominated by comparisons done exhaustively. As an alternative to the multi-biometric exhaustive search in the protected domain, this work extends the proof-of-concept of frequent binary patterns~\cite{OsorioRoig-IndexingDeepCancelableTemplates-IJCB-2022} to indexing multi-biometric cancelable references by employing strategies of biometric information fusion described in the Sect.~\ref{sec:fusion-works}. In a nutshell, the concept of frequent binary patterns is employed as a multi-biometric efficient binning scheme where each bin (\ie a single frequent binary pattern) is built by fusing $m$ representations from protected reference templates and allows for indexing them in a single biometric identification transaction. Fig.~\ref{fig:overview} presents a conceptual overview of the proposed scheme. The design of the multi-biometric binning scheme is template-protection-scheme and biometric characteristic-agnostic which makes it easy to work across different cancelable biometrics extracting binary representations. Sect.~\ref{sec:frequent-binary} provides details on the approach that computes frequent binary patterns, Sect.~\ref{sec:indexing-scheme} describes three strategies of information fusion that result in stable frequent binary patterns for indexing, Sect.~\ref{sec:retrieval-scheme} describes the retrieval process for each type of information fusion. Sect.~\ref{sec:workload-reduction-equation} discusses the obtained workload reduction.

\subsection{Frequent binary pattern extraction}
\label{sec:frequent-binary}

Frequent binary patterns can be defined in a general concept for the enrolment and retrieval processes, respectively. Formally, the frequent binary patterns can be extracted from a binary representation as follows: let $f \in \{0,1\}^n$ be a bit-string of size $n$ and $k < n$ a given frequent pattern length. A set of unique binary patterns $\mathbf{P}=\{p_1, \dots, p_{L}\}$, each of length $k$ can be computed over $f$ by sampling in a sliding window the consecutive $k$ bits starting from positions $[0, \ldots, n-k]$ with stride 1. In addition, let $\mathbf{O}=\{o_1, \dots, o_{L}\}$ be the set of occurrences of each $p_i \in \mathbf{P}$. Obviously, there is a direct relation between $\mathbf{O}$ and $\mathbf{P}$: for each $p_i \in \mathbf{P}$ there exists an $o_i \in \mathbf{O}$ which denotes the number of occurrences of $p_i$ in $f$. Therefore, for a general retrieval process, consider a function $\mathbf{FP}(\cdot)$ that extracts the set $\mathbf{P}$ ordered descending according to $\mathbf{O}$. 

 \subsection{Indexing multi-biometric frequent binary patterns}
\label{sec:indexing-scheme}

Conceptually, as mentioned above, frequent binary patterns can be extracted only from binary representations. Therefore, deciding which type of information to fuse from the protected references before or after extracting the patterns could impact the efficacy of the proposed binning scheme. Introducing known and simple fusion strategies (\eg concatenation) on intelligent and convenient steps increases the stability and the discriminative power of the procedure of frequent binary pattern extraction, thereby improving the overall results of the proposed system in terms of biometric performance and computational workload. 


Formally, let $\mathbf{R}_i=\{r^{1}_i, \dots, r^{m}_i\}$ be the set of data of the subject $i \in \{1,\dots,N\}$ in the enrolment database, where each $r^{j}_i$ denotes a protected binary reference associated with the  biometric characteristic $j\in \{1,\dots,m\}$. Given a fixed frequent binary pattern length of $k$ bits, the goal is to build a multi-biometric and efficient binning scheme over the base of stable frequent binary patterns successfully extracted on $\mathbf{R}_i$. For enrolment, this work considered the fusion strategies at two levels based on the concept of frequent binary patterns: feature and representation level. The former pipeline introduces the concatenation of protected binary references corresponding to different $m$ biometric characteristics. Here, the concatenation acts as doubling the feature dimension by keeping all the elements from the input features. The latter shows the fusion across the maximum binary patterns successfully mapped from individual protected binary references corresponding to $m$ biometric characteristics.

\begin{LaTeXdescription}

\item[Feature-level Indexing] each $r^{j}_i \in \mathbf{R}_i$ of size $d$ is concatenated with the remaining elements in $\mathbf{R}_i$ yielding a protected feature of size $d \cdot m$ bits. Let $\mathbf{B}_i=\Bigr[r^{1}_i\mathbin{\|}\dots\mathbin{\|}r^{m}_i\Bigr]$ be the concatenation of $m$ protected binary references of the $i$-th subject. $\mathbf{B}_i$ can be then mapped to an individual bin $b_i$ which is computed by $\max(\mathbf{FP}(\mathbf{B}_i))~\rightarrow~b_i$ given a fixed $k$, as explained in Sect.~\ref{sec:frequent-binary}. That is, the set of data subjects is indexed with at most $2^k$ bins. 

\item[Representation-level Indexing] each $r^{j}_i \in \mathbf{R}_i$ can be independently mapped by the function $\mathbf{FP}(r^{j}_i) ~\rightarrow~\mathbf{P}^{j}_i$, resulting in at most $2^k$ patterns. In this context, two fusion approaches are considered:

\begin{enumerate}
    \item Ranked-codes: $\max(\mathbf{P}^{j}_i)~\rightarrow~b_i$ a single binary pattern resulting in the most ranked frequent binary pattern extracted from the set $\{\mathbf{P}^{j}_i\}_{j=1}^m$  is considered as a stable bin for indexing.
    \item XOR-codes: the bin $b_i$ is constructed from the bitwise $\mathbf{XOR}$ operation between the binary patterns with the maximum occurrence in each $\mathbf{P}^{j}_i$ with $~1~\leq~j~\leq~m$, \ie $\mathbf{XOR}(\max(\mathbf{P}^{j}_i))~\rightarrow~b_i$.  
    
\end{enumerate}

\end{LaTeXdescription}

\subsection{Multi-biometric retrieval by fusion strategy}
\label{sec:retrieval-scheme}
As explained in Sect.~\ref{sec:frequent-binary}, for a general retrieval process, frequent binary patterns are extracted preserving their order of occurrence. It is expected that the pattern with the highest occurrence provides a better chance to find the correct candidate subject than patterns leading with low occurrence as showcased in~\cite{OsorioRoig-IndexingDeepCancelableTemplates-IJCB-2022}. In a retrieval step, this parameter (\ie pattern with the highest occurrence) would be estimated on an incremental search for those $p$ patterns with the highest occurrence. For a concrete example, consider $2^3$ patterns: $\mathbf{P} = \{p_1, p_2, \ldots, p_8 \}$, extracted by the function $\mathbf{FP}(\cdot)$ given $k=3$. A threshold $t$ with $1\leq t \leq 2^k$ is determined on $\mathbf{P}$ and represents the maximum number of bins that can be visited for a biometric probe. Note that this parameter ($t$) can easily be controlled by the binning scheme and is independent of the retrieval strategy employed. Also, extracted patterns can only take advantage of their orderings which can be influenced by the retrieval strategy employed (\eg type of fusion). In this regard, all proposed retrieval strategies employ a score-level fusion in a multi-biometric identification transaction once a corresponding bin is determined. In particular, a sum-rule fusion is applied among normalised similarity scores computed from each biometric characteristic. This type of fusion has been utilised in multi-biometric indexing schemes (\eg~\cite{Drozdowski-CascadingFiltering-TBIOM-2020}) and has also contributed to very good biometric performance in general (see \cite{Jain-guidelines-2015} and ISO/IEC TR 24722~\cite{ISO-IEC-24722-TR-Fusion-150216}).


In this work, three retrieval strategies, one for each type of information fusion are proposed. Firstly, we consider the fact that a binning scheme can be created using one of the strategies described in Sect.~\ref{sec:indexing-scheme}. In a retrieval scenario, let $\mathbf{Z}=\{z_1,\dots,z_m\}$ be the set of protected biometric templates for a probe subject, where each $z_j$ denotes a binary representation for each of the $m$ biometric characteristics. The key idea is that the proposed retrieval schemes offer different orderings and representations of the extracted frequent binary patterns. Subsequently, a parameter $t$ can be empirically computed in a multi-biometric identification transaction (see Sect.~\ref{sec:results-multi-modality}), thereby reducing the system workload while preserving a trade-off between biometric performance, efficiency and privacy.  

\begin{LaTeXdescription}
\item[Feature-level Retrieval] we follow a similar idea to that of feature-level indexing, as explained above in Sect.~\ref{sec:indexing-scheme}. Let $\mathbf{B}=\Bigr[z_1\mathbin{\|}\dots\mathbin{\|}z_{m}\Bigr]$ be the concatenation of all $ z_j \in \mathbf{Z}$ and a fixed $k$, the retrieval strategy searches the database for bins belonging to the ordered set $\mathbf{P}~\leftarrow~\mathbf{FP}(\mathbf{B})$. The final candidate list is therefore composed of the identities associated with the retrieved bins in $\mathbf{P}$.

\item[Representation-level Retrieval] in contrast to the feature-level based retrieval, this retrieval pipeline allows searching a $t$ by handling the binary patterns extracted per biometric characteristic. Said patterns are computed as follows:

\begin{enumerate}
    \item Ranked-codes: the database is searched for the highest ranked binary patterns of each $\mathbf{P}_j~\leftarrow~\mathbf{FP}(z_j)$ and the identities associated with those existing patterns make up the final candidate list.
    \item XOR-codes: the database is searched for those binary patterns resulting from the bitwise $\mathbf{XOR}$ operation among all possible pairs of binary patterns that belong to different $\mathbf{P}_j$. Note that the bitwise $\mathbf{XOR}$ operations are computed over at most $m~\cdot~2^k$ number of pattern pairings that can be constructed from $\{\mathbf{P}_j\}_{j=1}^m$.
\end{enumerate}

\end{LaTeXdescription}

\begin{figure}[!t]
    \centering
        \includegraphics[width=0.80\linewidth]{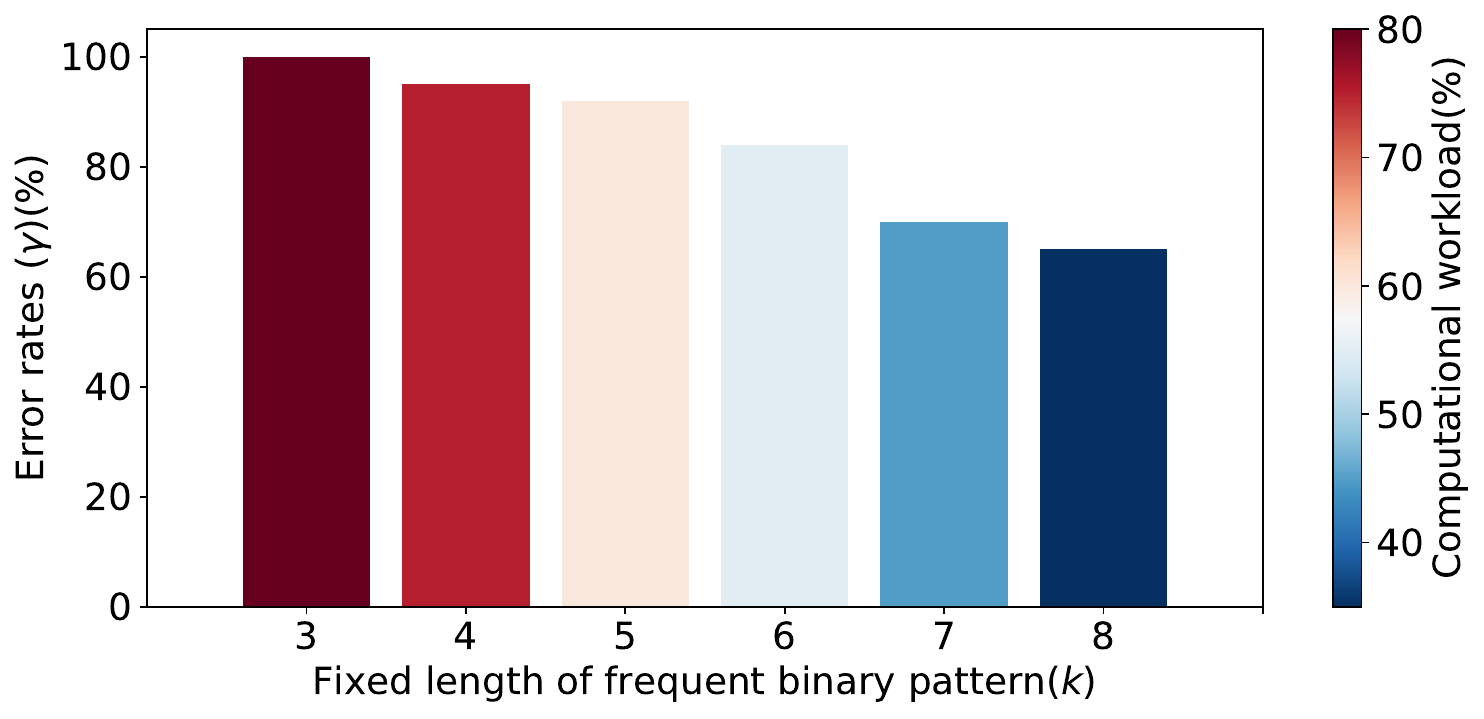}  
    \caption{Effect of $k$ on the trade-off between the biometric performance ($\gamma$) and the computational workload ($W_{proposed}$). Consider that a biometric performance can be computed in any biometric identification transaction (\eg $\gamma \rightarrow$ hit-rate).  }
    \label{fig:k-trade-offs}
\end{figure}

\begin{figure}[!t]
    \centering
        \includegraphics[width=0.80\linewidth]{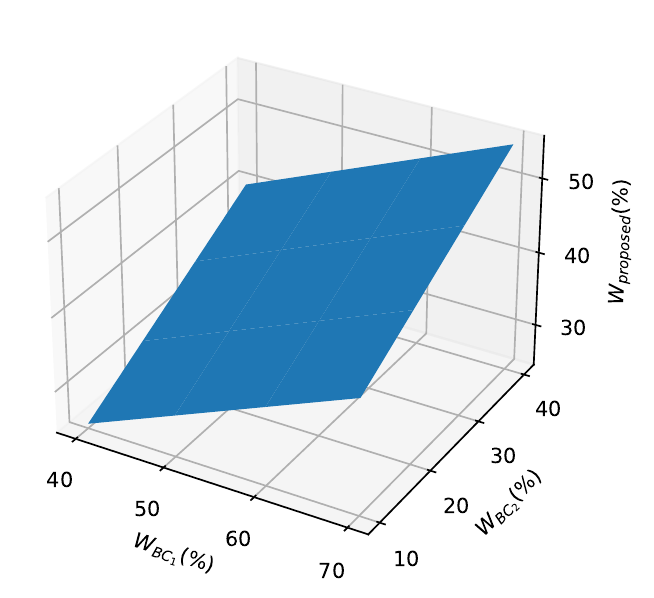}  
    \caption{Relation between the overall computational workload (\ie $W_{proposed}$) of the multi-biometric approach with respect to the individual workload by type of biometric characteristics involved (\ie $W_{BC_{1}}$ and $W_{BC_{2}}$). }

    
    \label{fig:workload-two-modalities}
\end{figure}

\subsection{Computational workload reduction}
\label{sec:workload-reduction-equation}

Our design, which is agnostic with respect to the type of biometric characteristics and cancelable schemes allows searching different biometric characteristics in a single biometric transaction. Therefore, the number of bins visited as well as the number of protected templates stored at each bin is expected to be the same per biometric characteristic. To that end, as mentioned in Sect.~\ref{sec:retrieval-scheme}, a threshold $t$ may be defined across $m$ types of biometric characteristics used. Although this parameter is easily managed by the multi-biometric binning scheme, a computational workload cost may be noticed depending on the biometric characteristics involved (\eg face and iris, or face and fingerprint), the workload of the individual biometric characteristics, and the strategy of fusion used for retrieval and indexing, respectively. 

The computational workload $W_{proposed}$ of an identification transaction (measured in terms of the number of necessary template comparisons) in the proposed scheme, can be expressed as follows:

\begin{equation}
\label{eq:workload}
W_{proposed} = \sum_{i=1}^{t}\vert b_{i}\vert \cdot m,
\end{equation}

\noindent where $1 \leq t \leq 2^{k}$ denotes a threshold for the maximum number of bins or frequent binary patterns visited in a retrieval step for a fixed $k$ and $\vert b_{i}\vert$ the number of protected templates associated with the biometric characteristics $m$ involved.  Note that $k$ implicitly is included in the Eq.~\ref{eq:workload} describing a fixed length in the search for frequent binary patterns (Sect.~\ref{sec:frequent-binary}), and is expected to have an effect on the computational workload along with the biometric performance (see Sect.~\ref{sec:results}). This trend is shown theoretically in Fig.~\ref{fig:k-trade-offs}. According to Fig.~\ref{fig:k-trade-offs}, it should be observed that larger $k$ appears to provide a discriminating effect on the built bins, reducing the number of protected templates stored within a bin and thus the overall computational workload. However, some deterioration in biometric performance is observed while maintaining a low workload. Additionally, Eq.~\ref{eq:workload} shows the relation between the computation of the overall computational workload and the types of biometric characteristics involved in the multi-biometric binning scheme. Fig.~\ref{fig:workload-two-modalities} theoretically visualizes said relation for \eg two biometric characteristics involved ($BC_{1}$ and $BC_{2}$) in a range of computed individual workloads. Note that individual workloads can be computed on the proposed scheme using a single biometric characteristic (\ie uni-modal biometric system). As observed, the overall computational workload appears to be directly proportional to the individual workloads corresponding to each biometric characteristic: $W_{proposed}$ increases with the workloads of the types of biometric characteristics involved. Also, this trend allows some biometric characteristics to take advantage of the unbalanced workloads among individual biometric characteristics, \eg $BC_{1}$ over $BC_{2}$ in this case. However, an improvement in overall biometric performance is expected to be achieved, albeit with a slight increase in the overall computational workload.

In summary, the key idea behind Eq.~\ref{eq:workload} is to reduce the computational workload dominated by the cost of comparisons carried out exhaustively. Hence, it is expected that $W_{proposed} \ll W_{baseline}$, reducing the penetration rate in the search. An upper bound of $W_{proposed}$ in Eq.~\ref{eq:workload} is reached, when the system retrieves all bins, resulting in an exhaustive search. In contrast, the best case is when the biometric probe is in the first bin retrieved (\ie $t = 1$) and this contains the fewest number of protected multi-biometric templates.  

\subsection{Privacy protection}
\label{sec:privacyprotection}
In the proposed system, indexing is performed on protected biometric templates. Obtaining indexes from these cancelable binary templates offers the advantage that the privacy protection of the underlying cancelable scheme is not impaired by the indexing scheme. Recently, it has been shown that indexing methods can leak sensitive information, in particular, if additional indexing data is extracted from unprotected biometric templates \cite{Bauspiess-HEBI-IJCB-2023}. In contrast, the proposed scheme extracts the indexing data from the protected templates. This means the privacy protection capability of the used cancelable biometrics scheme is maintained which is a major advantage of the presented indexing method. Precisely, requirements of irreversibility, unlinkability, and renewability are retained from the cancelable biometric scheme. Obviously, the frequent binary patterns extracted from the protected templates do not comprise any additional sensitive information that could be leveraged by an attacker. Due to this reason, the experiments will only focus on biometric performance. For privacy protection analysis the interested reader is referred to the corresponding publication of used cancelable schemes.

\begin{table}[!t]
	\scriptsize
	\begin{center}
	\caption{Summary of the datasets used in identification experiments.}
	\label{tab:description-db}
    \begin{adjustbox}{max width=\textwidth}
			  \begin{threeparttable}

	 \begin{tabular}{c c c c} \toprule
		 \textbf{Biometric characteristic} & \textbf{Dataset} & \textbf{\#Instances} &  \textbf{\#Samples} \\ \midrule
                Face&LFW~\cite{Huang-wild-2007}&1,120&4,126 \\ \cmidrule{1-4}
                Fingerprint&MCYT~\cite{Ortega-Garcia-MCYT-2003}&1,120&13,440\\ \cmidrule{1-4}
                \multirow{ 2}{*}{Iris}&CASIA-Iris-Thousand~\cite{casia-2004} &916&3,369 \\
                    &BioSecure~\cite{Ortega-multiscenario-2009}&204& 757\\
        
   	\bottomrule
	\end{tabular}

	\end{threeparttable}
	\end{adjustbox}
    \end{center}
\end{table}

\section{Experimental setup}
\label{sec:experimental-setup}
This section describes a detailed setup of the experiments conducted on privacy-preserving multi-biometric indexing. Sect~\ref{sec:dataset} describes the datasets together with the different biometric characteristics employed in this investigation, Sect.~\ref{sec:deep-templates} provides details of the extraction process of deep templates, Sect.~\ref{sec:cancelable-schemes} details the cancelable template protection schemes, while Sect.~\ref{sec:evaluation-metrics} provides the metrics for the evaluation of the proposed system.

\begin{figure}[!t]
\centering
\subfloat[LFW]{\includegraphics[width=0.45\linewidth]{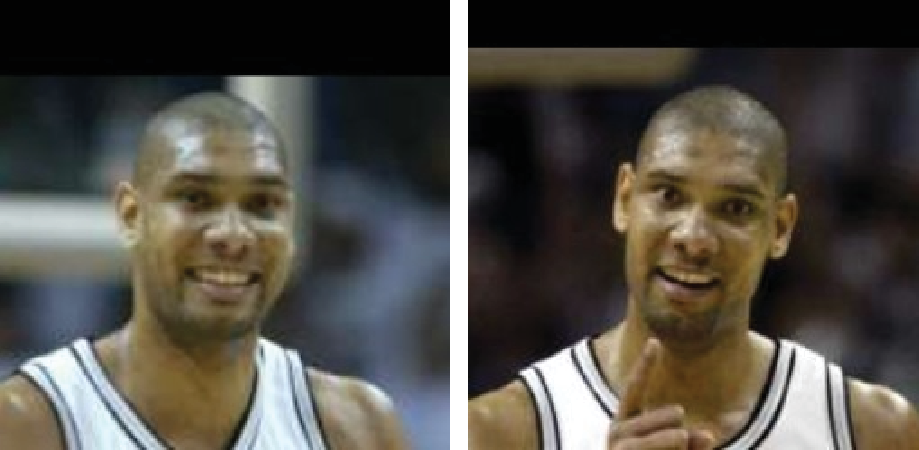}}%
\hspace{0.30cm}
\subfloat[MCYT(optical)]{\includegraphics[width=0.45\linewidth]{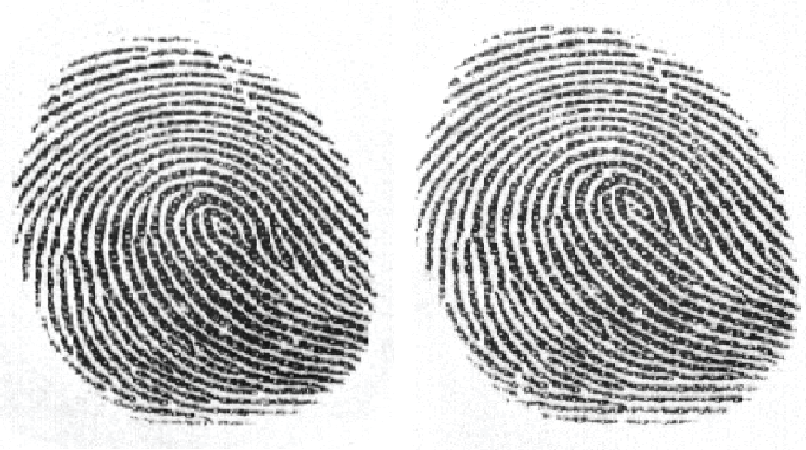}}%
\\
\subfloat[CASIA-Iris-Thousand]{\includegraphics[width=0.45\linewidth]{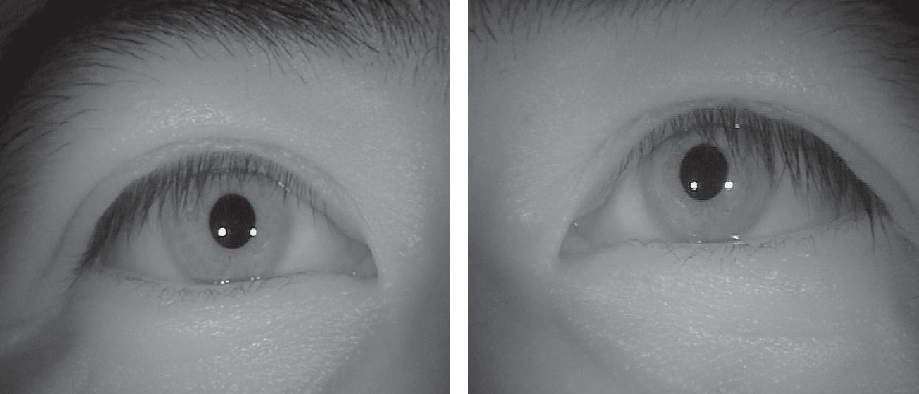}}%
 \hspace{0.30cm}
\subfloat[BioSecure]{\includegraphics[width=0.45\linewidth]{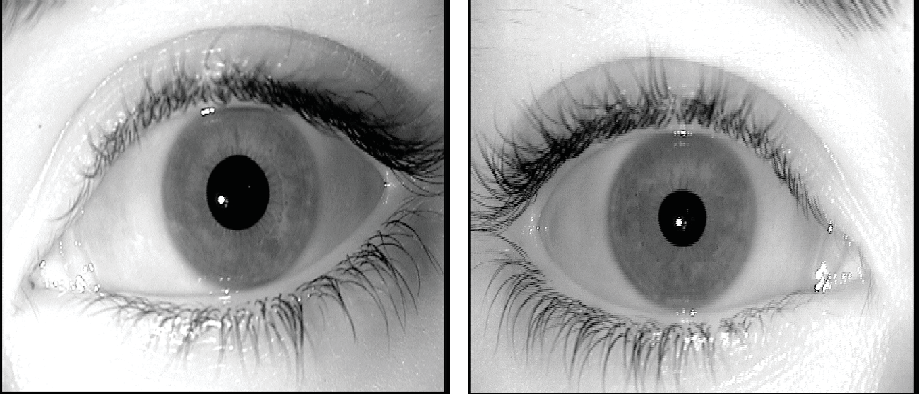}}%
\caption{Example images from the selected databases.}
\label{fig:databases-examples}
\end{figure}

\subsection{Databases}
\label{sec:dataset}
For biometric identification experiments where the workload reduction across indexing schemes is analysed, large-scale databases should be considered. Since large-scale databases are not available to researchers, we created a composite database using selected biometric characteristics. This type of database allows operational systems to work independently of the biometric characteristics and their feature representations. A similar concept was utilised in~\cite{Drozdowski-CascadingFiltering-TBIOM-2020}. Tab.~\ref{tab:description-db} shows an overview of the databases used in terms of the number of instances and samples. Note that we selected the three most common types of biometric characteristics, \ie face, fingerprint, and iris for our research. Details of the selected biometric characteristics and their databases are described as follows:

\begin{LaTeXdescription}
\item \textit{Face}: LFW~\cite{Huang-wild-2007} database is focusing on the large-scale unconstrained face recognition problem. It comprises 13,233 face images captured in the wild from 5,749 subjects collected from the web where 1,680 subjects are represented with two or more images and 4,069 subjects are represented with a single sample. In our experiments, we used only 1,170 identities from the group containing more than one face sample. CR-FIQA(L)~\cite{Boutros-quality-crfiqa-2023} as a quality measure has been utilised as a filtering step for selecting the subset of identities with their corresponding samples. 

\item \textit{Fingerprint}: MCYT~\cite{Ortega-Garcia-MCYT-2003} database containing only fingerprint images captured with an optical capture device is used. This dataset contains all 10 fingers from 330 subjects and 12 samples for each finger, for a total of 39,600 samples. For the experimental protocol, each fingerprint is considered a different biometric instance and is therefore treated as a separate subject. In particular, 1,170 identities are filtered out by using the quality factor NFIQ 2.0~\cite{NFIQ-2021}. 

\item \textit{Iris}: a mixed iris database was designed to achieve a balanced number of identities with respect to the other biometric characteristics. CASIA-Iris-Thousand~\cite{casia-2004} database and BioSecure~\cite{Ortega-multiscenario-2009} database, both containing images captured in the near-infrared light spectrum, were used. The former contains 1,000 subjects with 2,000 instances, each one represented with 10 samples from each right and left eye. The latter comprises 210 subjects with 420 instances, each one containing 4 samples from each right and left eye. Similar to the fingerprint biometric characteristic, each iris instance was considered a separate identity. In our experiments, a mixed subset was constructed up to a total of 1,170 identities. Iris samples and instances were discarded taking into account different criteria: segmentation errors that led to a bad normalisation step, samples containing glasses, critical images where the visible iris area did not represent more than 70\% of the usable iris area, and other quality measures with less critical behaviour such as the iris-sclera, iris-pupil contrast, and the iris-pupil ratio. Note that all quality measures analysed were evaluated and interpreted according to ISO/IEC 29794-6~\cite{ISO-IEC-29794-6-IrisQuality-150629}. For the quality assessment of iris samples, an open-source software\footnote{https://github.com/mitre/biqt-iris} BIQT-Iris was utilised, which reports all the quality measures described in ISO/IEC 29794-6.
\end{LaTeXdescription}

Note that 50 identities of 1,170 per database are selected for the score normalisation process and the remaining 1,120 are selected for biometric identification experiments. It is worth noting that biometric identification scenarios are more challenging than verification scenarios as the chance of a false positive can easily increase with the number of comparisons~\cite{Daugman-BiometricDesignLandscapes-2000}. Thus, for the selection of identities per biometric characteristic, the correlation between those biometric samples that produce worse similarity scores (highest chances of false positive in a critical operational point) and the different quality measures were also analysed. It should be noted that quality metrics were evaluated in order to keep only those samples with the best quality. To sum up, it is reasonable that biometric identification systems in real applications may operate with samples that provide acceptable quality in accordance with evaluation standards. Even more when biometric template protection and indexing schemes are employed. For the evaluations of the proposed multi-biometric systems, a database merged with the identities selected independently for each biometric characteristic (face, fingerprint, and iris) is constructed. Fig.~\ref{fig:databases-examples} shows some example images from the databases selected per biometric characteristic.

\subsection{Deep templates and pre-processing}
\label{sec:deep-templates}

For the experimental analysis, embeddings extracted by the current state-of-the-art DNN-based recognition systems per biometric characteristic are considered. All embeddings utilised consist of 512 floating-point values. Note that the features extracted per biometric characteristic are balanced in terms of the number of dimensions which allows the indexing scheme to produce the same chances of binary pattern search when they are fused at \eg the feature level. Details of the extraction process and pre-processing per biometric characteristic are described as follows:

\begin{LaTeXdescription}

    \item  \textit{Face}: ElasticFace~\cite{Boutros-elasticface-2022} represents a state-of-the-art face recognition system. Features are extracted from the pre-trained model ElasticFace made available by the authors~\footnote{https://github.com/fdbtrs/ElasticFace}.

    \item  \textit{Fingerprint}: deep fingerprint fixed-length representations are extracted from the open-source software introduced in~\cite{Rohwedder-FixedLengthFingerprintDNN-BIOSIG-2023}. Note that fingerprint embeddings are extracted by using the training on the texture branch.
    
    \item  \textit{Iris}: a deep iris representation extractor presented in~\cite{Boutros-deep-iris-2022} was used to extract iris embeddings. To that end, the approach proposed by~\cite{Boutros-deep-iris-2022} was trained on subsets of the CASIA-Iris-Thousand~\cite{casia-2004} and BioSecure~\cite{Ortega-multiscenario-2009} databases, respectively, from scratch. Note that those instances selected for training were not included in the set of instances for testing that contributed to the biometric identification experiments in this paper. Specifically, for the set of training, 200 instances~\footnote{those instances containing more than 3 samples} and 818 instances~\footnote{those instances containing more than 5 samples} from BioSecure~\cite{Ortega-multiscenario-2009} and CASIA-Iris-Thousand~\cite{casia-2004}, respectively, were selected randomly. ResNet50~\cite{Duta-backbone-resnet-2021} architecture is used as the backbone to extract iris feature representations and ArcFace~\cite{Deng-arcface-2019} loss function was employed in the training process. Iris images were pre-processed with the traditional approaches: iris segmentation was applied by using the Viterbi~\cite{Sutra-viterbi-2012} algorithm available in the open-source OSIRIS~\cite{Othman-osiris-2016}, iris textures were normalised according to the rubbersheet model~\cite{Daugman-iris-2009}, and subsequently, enhanced by applying Contrast Limited Adaptive Histogram Equalization(CLAHE)~\cite{Zuiderveld-clahe-1994}. 
\end{LaTeXdescription}


To sum up, it is important to note that any specific pre-processing like alignment, or type of input to the DNN was considered as described in their corresponding articles of reference. Furthermore, original embeddings extracted per biometric characteristic are converted to 512 binary-values feature vectors (\ie unprotected baseline system) by using a simple sign function with threshold 0. This type of representation is feasible for the design of the proposed scheme enabling the one-to-one comparison via hamming distance.

\subsection{Cancelable schemes}
\label{sec:cancelable-schemes}

Biometric template protection approaches representing the current state-of-the-art for cancelable schemes have been used in these experiments. In particular, the so-called BioHashing~\cite{Jin-BioHashing-2004} and a single instance of the Locality Sensitive Hashing~\cite{Jin-index-of-hashing-2017} based on Index-of-Maximum Hashing with Gaussian Random Projection (IoM-GRP). The former yields output representations containing 512 binary-point values, while the latter comprises 512 integer-point values. In order to facilitate the design agnostic \wrt the output of the cancelable scheme (binary representation) prior to the application of the proposed indexing scheme, output representations of the IoM-GRP approach were binarised prior to the frequent binary pattern extraction process. To that end, each integer value is encoded in $n$ bits which are computed on a one-hot encoding by using the maximum number of Gaussian Random Projection vectors ($q$) for all the IoM-GRP integer representations. Finally, a binary representation with length $n \cdot m$ bits, where $m$ represents the number of Gaussian Random Matrices or length of the integer representation, can be obtained. In these experiments, we used $q=16$ and $m=512$ to obtain a binary vector of size 2,048 bits. Note that for the computation of the similarity score function for a single biometric identification transaction, this scheme employs its own comparator based on the number of collisions through integers. In particular, BioHashing~\cite{Jin-BioHashing-2004} employs hamming distance. Overall, all protected templates have been used on stolen-token scenarios where non-mated comparisons have access to the genuine users' secret key and use this key with the impostors' own deep features.




\subsection{Proposed system configurations}
\label{sec:config-proposal}

Biometric identification experiments including the exhaustive search, \ie baseline workload ($W_{baseline}$), and the proposed indexing scheme (\ie at the feature- and representation-level) were conducted using 10-fold cross-validation for closed-set and open-set scenarios, respectively. For each fold, two samples per instance are randomly selected, one for enrolment and the other for search. It should be noted that the same samples (same randomness) selected for enrolment and search for each fold are maintained across the configurations of the proposed indexing schemes. Note also that the proposed multi-biometric approach applies to all possible combinations of two types of biometric characteristics and to all three types of biometric characteristics together. Moreover, for the step of score normalisation, the Z-score method is utilised as done in~\cite{Drozdowski-CascadingFiltering-TBIOM-2020}, which uses the arithmetic mean and standard deviation of the scores data. 

\subsection{Evaluation metrics}
\label{sec:evaluation-metrics}

The experimental evaluation is conducted according to two key aspects which are considered using methods and metrics standardised from the ISO/IEC 19795-1:2021~\cite{ISO-IEC-19795-1-060401} and supported by others which are commonly reported in the scientific literature:

\begin{itemize}
    \item \textbf{Biometric performance}: for the closed-set scenario, the hit-rate (H-R), the proportion of subjects for which the corresponding subject identifier is in the subset of candidates retrieved by the proposed indexing scheme; for the open-set scenario, the detection error trade-off (DET) curves between the false negative identification rate (FNIR) and false positive identification rate (FPIR). 
        
    \item \textbf{Computational workload reduction}: average proportion of the total number of references that are retrieved per identification transaction (denoted $W$) compared to a baseline workload (\ie an exhaustive search). It is worth noting that $W$ is theoretically defined in Sect.~\ref{sec:workload-reduction-equation}. 
    
\end{itemize}

\section{Results and discussion}
\label{sec:results}
In this section, the experimental results are described. Firstly, in Sect.~\ref{sec:results-single-modality}, the proof-of-concept of \textit{frequent binary patterns} for indexing deep cancelable templates is empirically validated to work on a single-biometric characteristic: face, iris, and fingerprint, against a baseline workload (\ie exhaustive search). Subsequently, Sect.~\ref{sec:results-multi-modality} shows the results of indexing by combining different biometric characteristics at different levels. It is worth noting that all the figures (plots) utilise nomenclatures to refer to the different types of biometric characteristics: FA(face), FP(fingerprint), and IR(iris). Also, different nomenclatures to refer to different statistical data computed in closed-set scenarios have been employed: \#Comp: Average number of comparisons, Std\_comp: Standard deviation across the comparisons done per subject, \#Visited-patterns: Average number of binary patterns visited from the probe, Std\_bins\_v: Standard deviation across the bins visited per subject.

\subsection{Single-biometric characteristic}
\label{sec:results-single-modality}

\begin{table*}[!t]
	\begin{center}
	\caption{Closed-set scenario over single-biometric characteristic. The baseline represents the unprotected system. Bold values show a trade-off between biometric performance and computational workload reduction. The description of the nomenclature named in the columns is detailed below the table.}
	\label{tab:single-modality-frequent-patterns-indexing}

\begin{adjustbox}{max width=\linewidth}
    \begin{threeparttable}
    \begin{tabular}{ccccccccccc}
    \toprule
        \textbf{BC} & \textbf{Approach} & \textbf{k} & \textbf{\#Comb} & \textbf{\#Comp} & \textbf{Std\_comp} & \textbf{W$_u$(\%)} & \textbf{W$_l$(\%)} & \textbf{\#Visited-patterns } &\textbf{Std\_bins\_v}& \textbf{H-R} \\ \cmidrule{1-11}
           \multirow{6}{*}{Face}& \multirow{2}{*}{Baseline}                          & \textbf{5} & \textbf{32} & \textbf{451.1} & \textbf{282.18} & \textbf{72.75} & \textbf{44.75} & \textbf{13.48} & \textbf{9.72} & \textbf{100.0} \\ 
                                &                            & 8 & 256 & 378.25 & 199.97 & 57.36 & 37.53 & 92.97 & 55.03 & 64.53 \\ \cmidrule{2-11} 
                                &\multirow{2}{*}{BioHashing} & \textbf{5} & \textbf{32} & \textbf{473.97} & \textbf{292.14} & \textbf{76.00} & \textbf{47.02} & \textbf{14.68} & \textbf{9.82} & \textbf{100.0} \\ 
                                &                            & 8 & 256 & 383.66 & 201.97 & 58.1 & 38.06 & 94.96 & 54.46 & 64.29 \\ \cmidrule{2-11}
                                & \multirow{2}{*}{IoM-GRP}   & \textbf{7} & \textbf{128} & \textbf{439.47} & \textbf{292.45} & \textbf{72.61} & \textbf{43.60} & \textbf{54.45} & \textbf{39.02} & \textbf{99.99} \\ 
                                &                            & 8 & 256 & 444.35 & 288.28 & 72.68 & 44.08 & 110.46 & 76.06 & 99.65 \\ \cmidrule{1-11} 
 \multirow{6}{*}{Fingerprint}  & \multirow{2}{*}{Baseline}   & \textbf{6} & \textbf{64} & \textbf{228.09} & \textbf{217.60} & \textbf{44.22} & \textbf{22.63} & \textbf{12.15} & \textbf{14.12} & \textbf{99.95} \\ 
                                 &                                 & 8 & 256 & 188.73 & 186.33 & 37.21 & 18.72 & 40.29 & 47.8 & 91.68 \\ \cmidrule{2-11}
                                 & \multirow{2}{*}{Biohashing} & \textbf{7} & \textbf{128} & \textbf{137.03} & \textbf{169.66} & \textbf{30.43} & \textbf{13.59} & \textbf{11.58} & \textbf{17.53} & \textbf{99.99} \\ 
                                 &                             & 8 & 256 & 117.58 & 155.03 & 27.05 & 11.67 & 17.91 & 29.69 & 98.66 \\ \cmidrule{2-11}
                                 &  \multirow{2}{*}{IoM-GRP}   & \textbf{7} & \textbf{128} & \textbf{245.4} & \textbf{196.82} & \textbf{43.87} & \textbf{24.35} & \textbf{8.12} & \textbf{13.73} & \textbf{100.0} \\
                                 &                             & 8 & 256 & 164.38 & 169.29 & 33.1 & 16.31 & 14.16 & 26.11 & 98.99 \\ \cmidrule{1-11}

    \multirow{6}{*}{Iris}        &\multirow{2}{*}{Baseline}  & \textbf{5} & \textbf{32} & \textbf{402.93} & \textbf{278.39} & \textbf{67.59} & \textbf{39.97} & \textbf{12.13} & \textbf{9.38} & \textbf{100.00} \\ 
                                 &  & 8 & 256 & 333.69 & 210.48 & 53.98 & 33.1 & 81.19 & 56.13 & 71.96 \\ \cmidrule{2-11}

                                 &\multirow{2}{*}{BioHashing}  & \textbf{6} & \textbf{64} & \textbf{410.67} & \textbf{290.29} & \textbf{69.54} & \textbf{40.74} & \textbf{25.35} & \textbf{19.31} & \textbf{99.46} \\ 
                                 &  & 8 & 256 & 342.75 & 208.49 & 54.69 & 34.0 & 83.57 & 55.46 & 72.13 \\ \cmidrule{2-11}
                                 &\multirow{2}{*}{IoM-GRP}  & \textbf{7} & \textbf{128} & \textbf{395.23} & \textbf{288.21} & \textbf{67.80} & \textbf{39.21} & \textbf{47.88} & \textbf{38.1} & \textbf{100.0} \\ 
                                 &  & 8 & 256 & 390.27 & 289.37 & 67.42 & 38.72 & 95.7 & 76.1 & 98.60 \\ 
                            
    \bottomrule
	\end{tabular}
		\begin{tablenotes}
  \item $k$: Length of the frequent binary pattern, \#Comb: Number of possible combinations to be generated given a $k$, \#Comp: Average number of comparisons, Std\_comp: Standard deviation across the number of comparisons carried out per subject, W$_l$: Lower bound of the computational workload reduction estimated on the average number of comparisons computed per subject, W$_u$: Upper bound of the computational workload reduction, \#Visited-patterns: Average number of binary patterns visited from the probe, Std\_bins\_v: Standard deviation across the bins visited per subject, H-R: Hit-Rate.
  \end{tablenotes}
	\end{threeparttable}
	\end{adjustbox}

    \end{center}
\end{table*}

\begin{table}[!t]
	\begin{center}
	\caption{Open-set results over single-biometric characteristic.}
	\label{tab:stat-open-set-exhaustive-indexing}
    
\begin{adjustbox}{max width=\linewidth}
    \begin{tabular}{ccccccc}
    \toprule
        \textbf{BC} & \textbf{Approach} &\textbf{k} &\textbf{\#Bins}& \textbf{W(\%)}& \textbf{FPIR=0.01(\%)} & \textbf{FPIR=0.1(\%)}\\ \midrule

                \multirow{3}{*}{\shortstack[c]{Face \\ (exhaustive)}} & Baseline &-&-&100.00&21.07 &17.85 \\
                                      & Biohashing&-&-&100.00&33.91 &21.42 \\
                                      & IoM-GRP&-&-&100.00&15.44 &14.09 \\ \midrule
                 \multirow{3}{*}{\shortstack[c]{Face \\ (indexing)}} & Baseline&5&23&69.99&35.00&34.00 \\ 
                                      & Biohashing&5&25&75.52&36.21&33.15 \\ 
                                      & IoM-GRP&7&93&71.22&37.61&32.10 \\ \midrule
                \multirow{3}{*}{\shortstack[c]{Fingerprint \\ (exhaustive)}} & Baseline&-&-&100.00&20.06 &15.79 \\
                                      & BioHashing&-&-&100.00&44.43&36.80 \\
                                      & IoM-GRP&-&-&100.00&16.92 &13.47 \\ \midrule

                \multirow{3}{*}{\shortstack[c]{Fingerprint \\ (indexing)}}  & Baseline&6&26&43.30  &28.76&24.86 \\ 

                                      & BioHashing&7&29&32.23&39.57&31.21 \\ 

                                      &IoM-GRP&7&22&35.86&22.47 &19.85 \\ \midrule
                                                                                         
            \multirow{3}{*}{\shortstack[c]{Iris \\ (exhaustive)}} & Baseline&-&-&100.00&44.86 &37.80 \\

                                  &BioHashing &-&-&100.00&44.60 &32.52 \\

                                  &IoM-GRP &-&-&100.00&36.75 &32.69 \\  \midrule

            \multirow{3}{*}{\shortstack[c]{Iris \\ (indexing)}}  &Baseline&5&22&67.06&70.59&49.70 \\ 

                                  &BioHashing&6&45&69.04&74.42&49.42 \\ 

                                  & IoM-GRP&7&86&66.63 &53.77&47.03 \\
        
    \bottomrule
     \end{tabular}
     \end{adjustbox}
    \end{center}
\end{table}

Tab.~\ref{tab:single-modality-frequent-patterns-indexing} shows the effect of the length of the frequent pattern (k) in relation to the hit rate (H-R) and the system workload (W) empirically computed for a set of identification transactions over closed-set scenario. Note that k has been only shown for the best configuration and for a final value of k (\ie k=8). An extended overview for all k-combinations can be found in the supplementary material Tab.~\ref{tab:SM_single-modality-frequent-patterns-indexing}. In the context of the workload computation, two workload reductions representing a lower bound (W$_l$) and an upper bound (W$_u$) are estimated. The former considers the lowest number of comparisons equitably distributed among bins without considering their standard deviations, while the latter considers an increased workload taking into account the standard deviations. Note that for a realistic scenario (\eg open-set scenario), the overall workload would be limited to the upper limit of computational workload (see W$_u$ on Tab.~\ref{tab:single-modality-frequent-patterns-indexing}) that can be easily controlled by a fixed number of bins for a biometric identification transaction. In addition to the closed-set scenario evaluations, Tab.~\ref{tab:stat-open-set-exhaustive-indexing} shows open-set results for the best parameter configurations in Tab.~\ref{tab:single-modality-frequent-patterns-indexing}. Note that for this scenario, a fixed number of bins representing the number of bins visited (see \#Visited-patterns + Std\_bins\_v in Tab.~\ref{tab:single-modality-frequent-patterns-indexing}) is set for a set of biometric identification transactions. It should be noted that an exhaustive search represents the baseline workload ($W_{baseline}=100\%$).

Tab.~\ref{tab:single-modality-frequent-patterns-indexing} shows that the proof-of-concept of \emph{frequent binary patterns} for indexing deep cancelable templates outperforms the exhaustive search in terms of workload reduction across different biometric characteristics for two well-known biometric template protection schemes: BioHashing and IoM-GRP. In particular, the lowest workloads observed for W$_u$ are 30.43\% and 43.87\% for BioHashing and IoM-GRP, respectively, and are achieved by the fingerprint while maintaining a high hit rate ({99\%$\leq$H-R$\leq$100\%}). Then, a higher workload can be perceived for the face (\ie W$_u\geq 72\%$) and iris (\ie W$_u\geq 67\%$) on the same schemes.

    
Additionally, it can be observed that the workload is inversely proportional to the length of the frequent pattern (k): workload decreases as the length increases, while some H-R values are compromised. A similar trend is theoretically shown in Fig.~\ref{fig:k-trade-offs} (Sect.\ref{sec:workload-reduction-equation}). This observation is to be expected, as bins constructed from longer lengths are more discriminative and can reduce the number of candidates in a comparison step. Therefore, this type of binning design makes the indexing scheme highly dependent on the intra-class and inter-class variance of each biometric characteristic. 


    
Focusing on the open-set results in Tab.~\ref{tab:stat-open-set-exhaustive-indexing}, at a fixed number of bins (see column \#Bins in Tab.~\ref{tab:stat-open-set-exhaustive-indexing}), it can be observed that indexing schemes for individual biometric characteristics do not achieve similar biometric performances with respect to their corresponding exhaustive searches. However, their workload reductions are remarkable with respect to the baseline workload. Also, the proposed multi-biometric indexing scheme is expected to outperform the biometric performance of the retrieved individual biometric characteristics, while maintaining the overall workload of the system.

\subsection{Multi-biometric characteristics}
\label{sec:results-multi-modality}

\begin{table}[!t]
\centering
	\caption{Closed scenario results of the proposed multi-biometric indexing schemes on Bio-Hashing for the best parameter k.}
 \begin{adjustbox}{width=\columnwidth}
	\label{tab:concatenation-feature-closed}

\begin{adjustbox}{max width=\linewidth}
    \begin{threeparttable}
    \begin{tabular}{ccccccc}
    \toprule
    \textbf{Method}  & \textbf{Combination}  & \textbf{k} & \textbf{W$_u$(\%)} & \textbf{W$_l$(\%)} & \textbf{H-R} \\ \midrule
\multirow{4}{*}{Feature-concatenation}   
&
\multirow{1}{*}{Face-Fingerprint}          &6&57.63 & 31.79&100.00   \\ 
&                                          
\multirow{1}{*}{Iris-Fingerprint}          &6&  52.04 & 27.63& 100.00 \\ 
&
\multirow{1}{*}{Face-Iris}                 &6&71.35 & 43.04&100.00    \\ 
&
\multirow{1}{*}{Face-Fingerprint-Iris}     &6& 61.53 & 34.68&100.00     \\ \midrule

\multirow{4}{*}{Ranked-codes}   
&
\multirow{1}{*}{Face-Fingerprint}          &6&  55.60 & 31.03& 99.72  \\       
&
\multirow{1}{*}{Iris-Fingerprint}          &6& 51.91 & 28.50 &  99.73   \\ 
&
\multirow{1}{*}{Face-Iris}                 &6& 71.80 & 43.44 & 99.28   \\ 
&                                    
\multirow{1}{*}{Face-Fingerprint-Iris}     &6&60.78 & 34.70 & 99.43  \\ \midrule  

\multirow{4}{*}{XOR-codes}     
&
\multirow{1}{*}{Face-Fingerprint}          &5& 78.00 &  49.97 & 100.00    \\  
&
\multirow{1}{*}{Iris-Fingerprint}          &5&78.55 &48.64 & 100.00  \\  
&
\multirow{1}{*}{Face-Iris}                 &7&  78.19 & 49.30 & 100.00  \\  
&
\multirow{1}{*}{Face-Fingerprint-Iris}     &7& 78.77 &51.18 & 100.00  \\

\bottomrule
	\end{tabular}
	\end{threeparttable}
	\end{adjustbox}
\end{adjustbox}
\end{table}

While Sect.~\ref{sec:results-single-modality} validated the concept of \emph{frequent binary patterns} as a solution agnostic \wrt types of biometric characteristics (\ie face, iris, and fingerprint) and cancelable biometric template protection schemes (with binary representation), this section shows the evaluation of different fusion strategies proposed in Sect.~\ref{sec:system}. Note that the proposed schemes allow the retrieval of protected multi-biometric templates in a single biometric transaction. Initially, the multi-biometric indexing results in a closed-set scenario are depicted in Tab.~\ref{tab:concatenation-feature-closed} for the best k-combinations across the proposed indexing approaches for the BioHashing scheme. An extended overview of all k-combinations and protection schemes can be found in the supplementary material Tab.~\ref{tab:SM_concatenation-feature-closed} (feature-concatenation indexing), Tab.~\ref{tab:SM_ranked-code-closed} (most ranked-code indexing), and Tab.~\ref{tab:SM_closed_set_xor_codes} (XOR-code indexing). Similar to single-biometric characteristic (Sect.~\ref{sec:results-single-modality}), the closed-set scenario evaluation allows estimating a threshold in terms of bins visited (see column \#Visited-patterns on the Tab.~\ref{tab:concatenation-feature-closed} in supplementary materials or see column \#Bins in Tab.~\ref{tab:stat-open-set-biohashing-indexing-score-fusion}) which can then be set in further open-set scenario evaluations. Since the order of the combinations between the biometric characteristics does not affect the final workload of the system in the closed-set scenario, a single combination for two and three biometric characteristics is shown. 

As observed in Tab.~\ref{tab:concatenation-feature-closed}, the computational workload required by the proposed multi-biometric techniques increases slightly or greatly depending on the type of strategy and the level at which the data are merged. The proposed multi-biometric approaches based on the highest-ranked code and feature-concatenation indexing improve the overall workload of some independent biometric characteristics, while slightly increasing the individual workload of others. More specifically, this trend is generally observed when designing combinations of types of biometric characteristics centred on the fingerprint. In other words, the Face-Fingerprint combination results in an average workload of approximately W$_l$=$\sim$31\%, which represents 16 percentage points lower than the workload of the face as a single biometric characteristic (\ie W$_l$=$\sim$47\% in Tab.~\ref{tab:single-modality-frequent-patterns-indexing}) and approximately 18 percentage points more than the workload yielded individually by the fingerprint (\ie W$_l$=$\sim$13\% in Tab.~\ref{tab:single-modality-frequent-patterns-indexing}). Similar trends can be also observed for the combinations with Iris, \eg Face-Fingerprint and Face-Iris. The above observations have also been modelled theoretically in Fig.~\ref{fig:workload-two-modalities} and Sect.~\ref{sec:workload-reduction-equation}. We believe that these gaps or imbalances in terms of overall workloads across the biometric characteristic combinations are due to the fact that single-biometric characteristics (\eg face or fingerprint) may exhibit different biometric variances (intra- and inter-class). Note that some variations are nearly inevitable and specific for some biometric characteristics, \eg for fingerprint, environmental conditions during the sample acquisition process and for iris, distance and angle from the sensor.


\begin{table}[!t]
\centering
\caption{Open-set results over BioHashing across different indexing-schemes.}
	\begin{adjustbox}{max width=\columnwidth}
	\label{tab:stat-open-set-biohashing-indexing-score-fusion}
    \begin{tabular}{ccccccc}
    \toprule
                &\textbf{BC} & \textbf{k} &\textbf{\#Bins}& \textbf{W(\%)}& \textbf{FPIR=0.01(\%)} & \textbf{FPIR=0.1(\%)}\\ \midrule

                 \multirow{3}{*}{\rotatebox[origin=c]{90}{Single}}&Face(indexing)             & 5 & 25 &75.52 &36.21  &33.15 \\ 
                                        
                                            &Fingerprint(indexing)      &7&29&32.23&39.57&31.21 \\ 
                                               
                                                                                                                                         
                                            &Iris(indexing)             &6&45&69.04&74.42&49.42 \\ \midrule
                                        
            
             \multirow{12}{*}{\rotatebox[origin=c]{90}{Feature-concatenation}}&Face-Fingerprint   &6&32&53.98&23.60&21.78\\
                                                  &Fingerprint-Face     &6&32&53.91&25.46&23.17\\
                                                  &Iris-Fingerprint     &6&31&53.07&32.20&24.69\\ 
                                                  &Fingerprint-Iris     &6&31&53.11&29.92&26.81\\
                                                  &Face-Iris            &6&46&70.22&27.73&24.56\\
                                                  &Iris-Face            &6&46&70.27&26.82&23.70\\ 
                                                  &Face-Fingerprint-Iris  &6&38&61.61&19.81&19.36\\
                                                  &Face-Iris-Fingerprint  &6&38&61.63&20.02&19.32\\  
                                                  &Fingerprint-Face-Iris  &6&38&61.63&20.47&18.38\\
                                                  &Fingerprint-Iris-Face  &6&38&61.63&20.56&19.00\\
                                                  &Iris-Face-Fingerprint  &6&38&61.59&22.57&18.84\\
                                                  &Iris-Fingerprint-Face  &6&38&61.59&19.98&18.38\\ \midrule

           \multirow{12}{*}{\rotatebox[origin=c]{90}{Ranked-codes}}&        Face-Fingerprint        &6&32&56.02&20.37&19.73\\
                                                  &Fingerprint-Face       &6&32&56.02&20.37&19.73\\
                                                  &Iris-Fingerprint       &6&30&53.40&32.14&25.36\\ 
                                                  &Fingerprint-Iris       &6&30&53.40&32.14&25.36\\
                                                  &Face-Iris              &6&46&72.92&26.16&23.49\\
                                                  &Iris-Face              &6&46&72.92&26.16&23.49 \\ 
                                                  &Face-Fingerprint-Iris  &6&33&57.40&21.55&20.81\\
                                                  &Face-Iris-Fingerprint  &6&33&57.40&21.55&20.81 \\  
                                                  &Fingerprint-Face-Iris  &6&33&57.40&21.55&20.81\\
                                                  &Fingerprint-Iris-Face  &6&33&57.40&21.55&20.81\\
                                                  &Iris-Face-Fingerprint  &6&33&57.40&21.55&20.81\\
                                                  &Iris-Fingerprint-Face  &6&33&57.40&21.55&20.81 \\ \midrule

        \multirow{12}{*}{\rotatebox[origin=c]{90}{XOR-codes}}&              Face-Fingerprint        &5&25&78.05&28.11&24.53\\
                                                  &Fingerprint-Face       &5&25&78.05&28.11&24.53\\
                                                  &Iris-Fingerprint       &5&25&78.03&35.25&29.36\\ 
                                                  &Fingerprint-Iris       &5&25&78.03&35.25&29.36\\
                                                  &Face-Iris              &7&100&78.13&28.93&26.67\\
                                                  &Iris-Face              &7&100&78.13&28.93&26.67\\ 
                                                  &Face-Fingerprint-Iris  &7&63&81.55&26.24&22.38\\
                                                  &Face-Iris-Fingerprint  &7&63&81.55&26.24&22.38\\  
                                                  &Fingerprint-Face-Iris  &7&63&81.55&26.24&22.38\\
                                                  &Fingerprint-Iris-Face  &7&63&81.55&26.24&22.38\\
                                                  &Iris-Face-Fingerprint  &7&63&81.55&26.24&22.38\\
                                                  &Iris-Fingerprint-Face  &7&63&81.55&26.24&22.38\\ 
    \bottomrule
     \end{tabular}

    \end{adjustbox}
\end{table}

Subsequently, the evaluation of the open-set scenario is shown in Tab.~\ref{tab:stat-open-set-biohashing-indexing-score-fusion} across the proposed multi-biometric indexing approaches for BioHashing. For convenience, the evaluation of each of the individual biometric characteristics indexing systems is also presented. In these experiments, all possible combinations of types of biometric characteristics and orderings are analysed. The results of the other protection schemes can be found in the supplementary material Tab.~\ref{tab:SM_baseline-open-set-multi-bio} (Baseline) and Tab.~\ref{tab:SM_grp-open-set-multi-bio} (IoM-GRP).

\begin{figure*}[!t]
\centering
\subfloat[Feature-concatenation]{\includegraphics[width=0.25\linewidth]{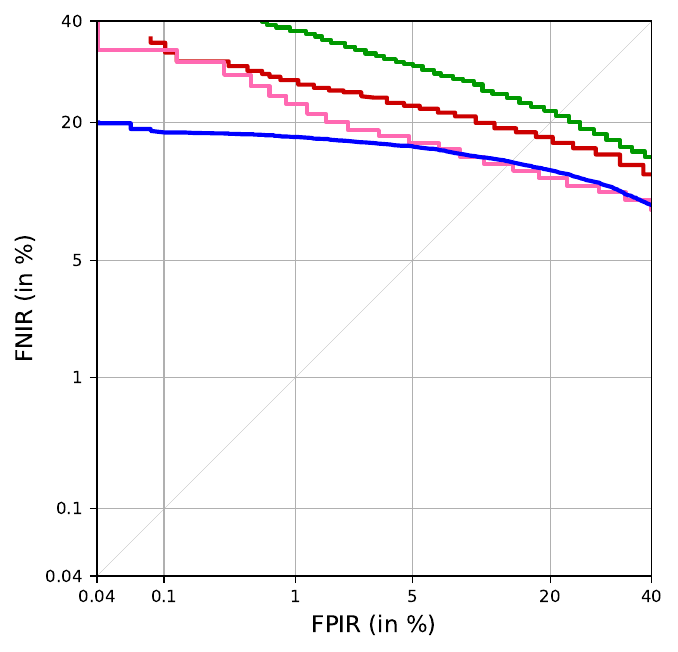}}%
\subfloat[Ranked-codes]{\includegraphics[width=0.25\linewidth]{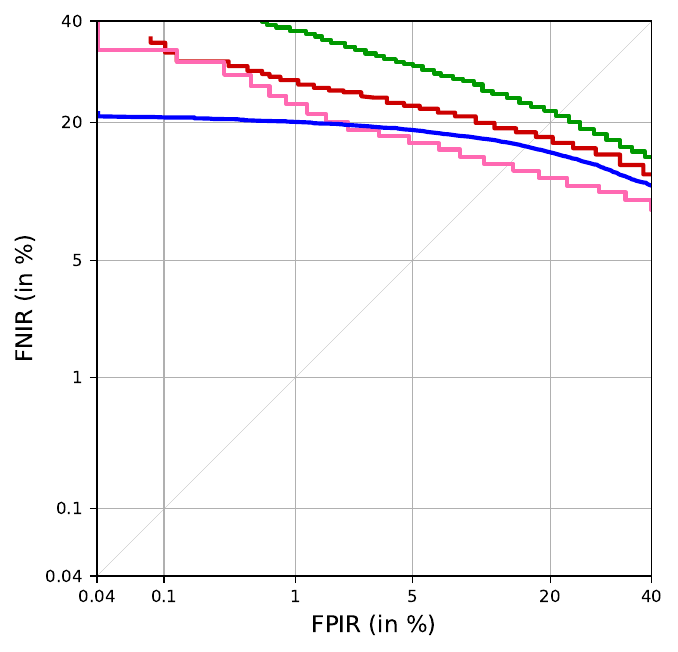}}%
\subfloat[XOR-codes]{\includegraphics[width=0.32\linewidth]{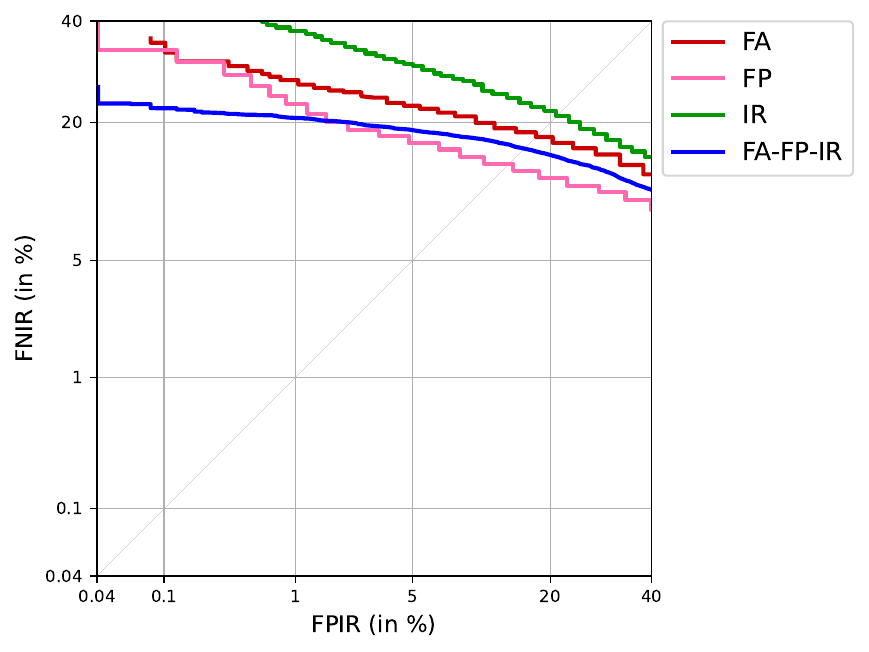}}%
\caption{Best results over open-set scenario are reported on the BioHashing for different multi-biometric approaches \wrt their uni-modal approaches.}
\label{fig:open-set-biohashing}
\end{figure*}

\begin{figure*}[!t]
\centering
\subfloat[Feature-concatenation]{\includegraphics[width=0.25\linewidth]{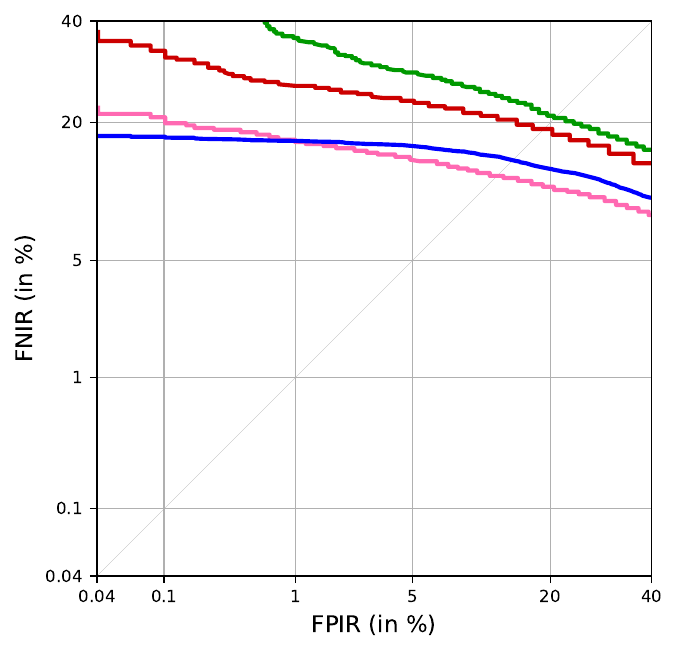}}%
\subfloat[Ranked-codes]{\includegraphics[width=0.25\linewidth]{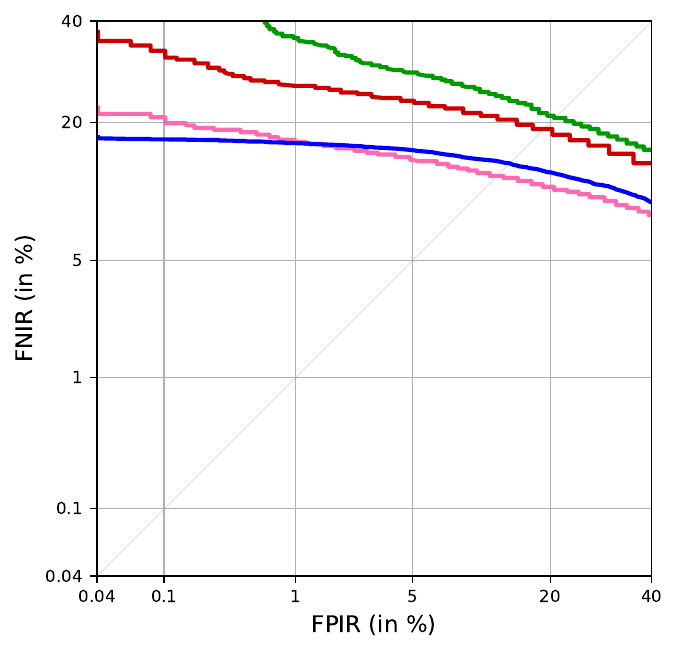}}%
\subfloat[XOR-codes]{\includegraphics[width=0.32\linewidth]{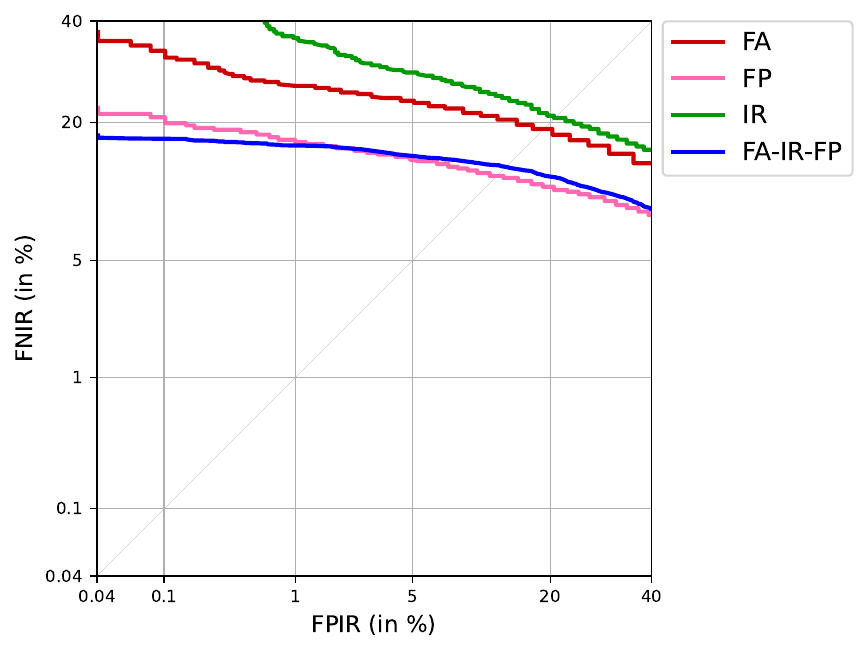}}%
\caption{Best results over open-set scenario are reported on the IoM-GRP for different multi-biometric approaches \wrt their uni-modal approaches.}
\label{fig:SM_open-set-grp}
\end{figure*}

Note that, on the one hand, the overall computational workload (W) of the different multi-biometric approaches proposed is not affected by the order of the biometric characteristics involved in the combination, \eg Face-Iris or Iris-Face. On the other hand, the workload (W) of the systems generally depends on the type of biometric characteristics used in the combination process, similar to what was observed for the closed-set scenario, \eg Face-Iris results in a higher W than the Fingerprint-Iris. Furthermore, the proposed multi-biometric schemes outperform single-biometric characteristic indexing pipelines in terms of biometric performance, while producing an approximate average W of the individual BCs. Note the imbalances in terms of W between multi-biometric and single-biometric characteristic systems. In particular, and depending on the multi-biometric strategy, the FNIR produced by the single-biometric characteristic approaches is reduced down to 19.81\% for high-security thresholds (\ie FPIR = 0.01\%). With regard to the above results, we also observe that the best trade-off between W and biometric performance is achieved by combining three biometric characteristics, \eg the ranked-codes approach results in a FNIR = 21.55\% at a FPIR = 0.01\%, which is approximately up to 53 percentage points less than the FNIR yielded \eg for Iris at the same operating point (FNIR~=~74.42\%). These performance trends are confirmed in Fig.~\ref{fig:open-set-biohashing}: The blue DET curves, representing the multi-biometric scheme merging three BCs, significantly outperform the remaining curves associated with the individual biometric characteristics for higher security thresholds. DET curves for other protection schemes can be found in the supplementary material Fig.~\ref{fig:SM_open-set-baseline} (Baseline) and Fig.~\ref{fig:SM_open-set-grp} (IoM-GRP).

\section{Conclusions}
\label{sec:conclusions}
A multi-biometric indexing scheme for binning and retrieving protected biometric templates is proposed. We show that the proposed approach is agnostic across biometric characteristics and cancelable biometric schemes. Focusing on unprotected biometric systems, some published works have reported results in terms of workload reduction that go beyond the approach presented~\cite{Drozdowski-WorkloadSurvey-IET-2019}. Nevertheless, most of these systems are custom-built for specific biometric systems and are not expected to be applicable to other systems. In contrast to these schemes, the proposed system can be used to merge different biometric characteristics, such as face, fingerprint and iris, while protecting the privacy of the subjects. Experimental evaluations compliant with the international metrics defined in the ISO/IEC 19795-1:2021~\cite{ISO-IEC-19795-1-060401} showed that a protected multi-biometric identification system can reduce the computational workload to approximately 57\% (indexing up to three types of biometric characteristics) and 53\% (indexing up to two types of biometric characteristics), while simultaneously improving the biometric performance at the high-security thresholds of a baseline biometric system.

\subsection{Acknowledgments}
\noindent This work has in part received funding from the European Union’s Horizon 2020  research and innovation programme under the Marie Skłodowska-Curie grant agreement No. 860813 - TReSPAsS-ETN and the German Federal Ministry of Education and Research and the Hessen State Ministry for Higher Education, Research and the Arts within their joint support of the National Research Center for Applied Cybersecurity ATHENE.

\section{Supplementary Material}

\begin{table*}[!t]
\centering
   \caption{Closed-set scenario over single-biometric characteristic. The baseline represents the unprotected system. Bold values show a trade-off between biometric performance and computational workload reduction. The description of the nomenclature named in the columns is detailed below the table.}
  \begin{adjustbox}{max width=0.8\linewidth}
   \label{tab:SM_single-modality-frequent-patterns-indexing}
    \begin{adjustbox}{max width=\linewidth}
    \begin{threeparttable}
    \begin{tabular}{ccccccccccc}
    \toprule
        \textbf{Biometric-characteristic} & \textbf{Approach} & \textbf{k} & \textbf{\#Comb} & \textbf{\#Comp} & \textbf{Std\_comp} & \textbf{W$_u$(\%)} & \textbf{W$_l$(\%)} & \textbf{\#Visited-patterns } &\textbf{Std\_bins\_v}& \textbf{H-R} \\ \cmidrule{1-11}
         \multirow{18}{*}{Face} &  \multirow{6}{*}{Baseline} & 3 & 8 & 509.62 & 278.66 & 78.2 & 50.56 & 3.65 & 2.46 & 100.0 \\
                                &                            & 4 & 16 & 477.87 & 282.08 & 75.39 & 47.41 & 7.06 & 4.89 & 100.0 \\ 
                                &                            & \textbf{5} & \textbf{32} & \textbf{451.1} & \textbf{282.18} & \textbf{72.75} & \textbf{44.75} & \textbf{13.48} & \textbf{9.72} & \textbf{100.0} \\ 
                                &                            & 6 & 64 & 445.32 & 288.08 & 73.76 & 44.18 & 27.09 & 19.56 & 99.41 \\ 
                                &                            & 7 & 128 & 437.83 & 278.07 & 71.02 & 43.44 & 53.69 & 37.39 & 92.67 \\ 
                                &                            & 8 & 256 & 378.25 & 199.97 & 57.36 & 37.53 & 92.97 & 55.03 & 64.53 \\ \cmidrule{2-11}
                                & \multirow{6}{*}{BioHashing}& 3 & 8 & 549.22 & 285.48 & 82.81 & 54.49 & 4.11 & 2.45 & 100.0 \\ 
                                &                            & 4 & 16 & 499.99 & 283.46 & 77.72 & 49.6 & 7.65 & 4.85 & 100.0 \\ 
                                &                            & \textbf{5} & \textbf{32} & \textbf{473.97} & \textbf{292.14} & \textbf{76.00} & \textbf{47.02} & \textbf{14.68} & \textbf{9.82} & \textbf{100.0} \\ 
                                &                            & 6 & 64 & 453.77 & 290.62 & 73.85 & 45.02 & 28.12 & 19.39 & 98.44 \\ 
                                &                            & 7 & 128 & 444.67 & 281.37 & 72.03 & 44.11 & 55.12 & 37.68 & 91.17 \\ 
                                &                            & 8 & 256 & 383.66 & 201.97 & 58.1 & 38.06 & 94.96 & 54.46 & 64.29 \\ \cmidrule{2-11}
                                & \multirow{6}{*}{IoM-GRP}       & 3 & 8 & 529.04 & 282.45 & 80.5 & 52.48 & 3.77 & 2.54 & 100.0 \\ 
                                &                            & 4 & 16 & 482.69 & 285.9 & 76.25 & 47.89 & 7.12 & 4.98 & 100.0 \\ 
                                &                            & 5 & 32 & 462.92 & 290.43 & 74.74 & 45.92 & 13.97 & 9.93 & 100.0 \\ 
                                &                            & 6 & 64 & 450.3 & 289.12 & 73.36 & 44.67 & 27.6 & 19.53 & 100.0 \\ 
                                &                            & \textbf{7} & \textbf{128} & \textbf{439.47} & \textbf{292.45} & \textbf{72.61} & \textbf{43.60} & \textbf{54.45} & \textbf{39.02} & \textbf{99.99} \\ 
                                &                            & 8 & 256 & 444.35 & 288.28 & 72.68 & 44.08 & 110.46 & 76.06 & 99.65 \\ \cmidrule{1-11}
    \multirow{18}{*}{Fingerprint} & \multirow{6}{*}{Baseline} & 3 & 8 & 370.92 & 225.48 & 59.17 & 36.8 & 2.36 & 1.89 & 100.0 \\ 
                             &                                & 4 & 16 & 294.03 & 220.19 & 51.01 & 29.17 & 3.86 & 3.67 & 100.0 \\ 
                             &                                & 5 & 32 & 254.91 & 218.09 & 46.92 & 25.29 & 6.67 & 7.14 & 100.0 \\ 
                             &                                & \textbf{6} & \textbf{64} & \textbf{228.09} & \textbf{217.60} & \textbf{44.22} & \textbf{22.63} & \textbf{12.15} & \textbf{14.12} & \textbf{99.95} \\ 
                             &                                & 7 & 128 & 211.11 & 216.15 & 42.39 & 20.94 & 22.33 & 27.73 & 98.90 \\ 
                             &                                & 8 & 256 & 188.73 & 186.33 & 37.21 & 18.72 & 40.29 & 47.8 & 91.68 \\ \cmidrule{2-11}
                                 & \multirow{6}{*}{Biohashing} & 3 & 8 & 324.75 & 211.99 & 53.25 & 32.22 & 2.22 & 1.7 & 100.0 \\ 
                                 &                             & 4 & 16 & 248.4 & 200.44 & 44.53 & 24.64 & 3.16 & 3.04 & 100.0 \\ 
                                 &                             & 5 & 32 & 186.2 & 180.84 & 36.41 & 18.47 & 4.68 & 5.22 & 100.0 \\ 
                                 &                             & 6 & 64 & 150.85 & 168.64 & 31.70 & 14.97 & 7.08 & 9.18 & 99.87 \\ 
                                 &                             & \textbf{7} & \textbf{128} & \textbf{137.03} & \textbf{169.66} & \textbf{30.43} & \textbf{13.59} & \textbf{11.58} & \textbf{17.53} & \textbf{99.99} \\ 
                                 &                             & 8 & 256 & 117.58 & 155.03 & 27.05 & 11.67 & 17.91 & 29.69 & 98.66 \\ \cmidrule{2-11}
                                 & \multirow{6}{*}{IoM-GRP}        & 3 & 8 & 455.66 & 226.07 & 67.63 & 45.2 & 1.81 & 1.52 & 100.0 \\ 
                                 &                             & 4 & 16 & 409.32 & 227.67 & 63.19 & 40.61 & 2.3 & 2.3 & 100.0 \\ 
                                 &                             & 5 & 32 & 358.59 & 222.95 & 57.69 & 35.57 & 3.07 & 3.77 & 100.0 \\
                                 &                             & 6 & 64 & 331.95 & 217.85 & 54.54 & 32.93 & 4.72 & 6.92 & 100.0 \\ 
                                 &                             & \textbf{7} & \textbf{128} & \textbf{245.4} & \textbf{196.82} & \textbf{43.87} & \textbf{24.35} & \textbf{8.12} & \textbf{13.73} & \textbf{100.0} \\
                                 &                             & 8 & 256 & 164.38 & 169.29 & 33.1 & 16.31 & 14.16 & 26.11 & 98.99 \\ \cmidrule{1-11}
        \multirow{18}{*}{Iris} & \multirow{6}{*}{Baseline} & 3 & 8 & 463.59 & 267.48 & 72.53 & 45.99 & 3.35 & 2.35 & 100.0 \\
                                 &  & 4 & 16 & 424.81 & 276.47 & 69.57 & 42.14 & 6.33 & 4.72 & 100.0 \\ 
                                 &  & \textbf{5} & \textbf{32} & \textbf{402.93} & \textbf{278.39} & \textbf{67.59} & \textbf{39.97} & \textbf{12.13} & \textbf{9.38} & \textbf{100.00} \\ 
                                 &  & 6 & 64 & 393.71 & 285.35 & 67.37 & 39.06 & 24.01 & 19.04 & 98.60 \\ 
                                 &  & 7 & 128 & 388.16 & 276.7 & 65.96 & 38.51 & 47.34 & 36.97 & 93.87 \\ 
                                 &  & 8 & 256 & 333.69 & 210.48 & 53.98 & 33.1 & 81.19 & 56.13 & 71.96 \\ \cmidrule{2-11}
                                 & \multirow{6}{*}{BioHashing} & 3 & 8 & 501.23 & 285.26 & 78.03 & 49.72 & 3.78 & 2.44 & 100.0 \\ 
                                 &  & 4 & 16 & 455.62 & 283.04 & 73.28 & 45.2 & 6.96 & 4.79 & 100.0 \\ 
                                 &  & 5 & 32 & 421.17 & 289.29 & 70.48 & 41.78 & 12.98 & 9.74 & 100.0 \\ 
                                 &  & \textbf{6} & \textbf{64} & \textbf{410.67} & \textbf{290.29} & \textbf{69.54} & \textbf{40.74} & \textbf{25.35} & \textbf{19.31} & \textbf{99.46} \\ 
                                 &  & 7 & 128 & 398.52 & 278.75 & 67.19 & 39.54 & 49.12 & 36.98 & 94.05 \\ 
                                 &  & 8 & 256 & 342.75 & 208.49 & 54.69 & 34.0 & 83.57 & 55.46 & 72.13 \\ \cmidrule{2-11}
                                 & \multirow{6}{*}{IoM-GRP} & 3 & 8 & 494.18 & 267.41 & 75.56 & 49.03 & 3.4 & 2.41 & 100.0 \\ 
                                 &  & 4 & 16 & 440.1 & 274.98 & 70.94 & 43.66 & 6.18 & 4.8 & 100.0 \\ 
                                 &  & 5 & 32 & 417.2 & 280.47 & 69.21 & 41.39 & 12.0 & 9.63 & 100.0 \\ 
                                 &  & 6 & 64 & 393.25 & 282.52 & 67.04 & 39.01 & 23.47 & 18.94 & 100.0 \\ 
                                 &  & \textbf{7} & \textbf{128} & \textbf{395.23} & \textbf{288.21} & \textbf{67.80} & \textbf{39.21} & \textbf{47.88} & \textbf{38.1} & \textbf{100.0} \\ 
                                 &  & 8 & 256 & 390.27 & 289.37 & 67.42 & 38.72 & 95.7 & 76.1 & 98.60 \\ 
                            
    \bottomrule
	\end{tabular}
		\begin{tablenotes}
  \item $k$: Length of the frequent binary pattern, \#Comb: Number of possible combinations to be generated given a $k$, \#Comp: Average number of comparisons, Std\_comp: Standard deviation across the number of comparisons carried out per subject, W$_l$: Lower bound of the computational workload reduction estimated on the average number of comparisons computed per subject, W$_u$: Upper bound of the computational workload reduction, \#Visited-patterns: Average number of binary patterns visited from the probe, Std\_bins\_v: Standard deviation across the bins visited per subject, H-R: Hit-Rate.
  \end{tablenotes}
	\end{threeparttable}
	\end{adjustbox}

    \end{adjustbox}
\end{table*}

\begin{table*}[!t]
\centering
	\caption{Closed scenario over multi-biometric characteristics. Results are shown for the feature-concatenation indexing.}
 \begin{adjustbox}{max width=0.8\linewidth}
    \label{tab:SM_concatenation-feature-closed}
    \begin{adjustbox}{max width=\linewidth}
    \begin{threeparttable}
    \begin{tabular}{ccccccccccc}
    \toprule
        \textbf{Combination}& \textbf{Approach} & \textbf{k} & \textbf{\#Comb} & \textbf{\#Comp} & \textbf{Std\_comp} & \textbf{W$_u$(\%)} & \textbf{W$_l$(\%)} & \textbf{\#Visited-patterns} &\textbf{Std\_bins\_v}& \textbf{H-R} \\ \cmidrule{1-11}
         \multirow{6}{*}{Face-Fingerprint}&\multirow{24}{*}{Baseline}&3&8 &452.90 & 258.01&70.53& 44.93&3.03&2.25&100.00 \\ 
                                          &                         &4&16&391.96& 262.90&64.97&38.88&5.45&4.44&100.00 \\
                                          &                         &5&32&356.03&260.67&61.18&35.32&9.99&8.78&100.00 \\
                                          &                         &6&64&336.83&261.94&59.40&33.42&19.25&17.54&100.00 \\
                                          &                         &7&128&317.98&262.64&57.60&31.55&36.58&34.71&98.75 \\
                                          &                         &8&256&306.90&258.46&56.07&30.45&70.91& 67.75& 96.86 \\ \cmidrule{3-11}

       \multirow{6}{*}{Iris-Fingerprint}  &                         &3&8&430.08&255.90&68.05&42.67&2.95&2.21&100.00 \\
                                          &                         &4&16&359.25&249.51&60.39&35.64&5.01&4.22&100.00 \\
                                          &                         &5&32&320.58&247.33&56.34&31.80&8.99&8.31&100.00 \\ 
                                          &                         &6&64&305.82&250.85& 55.23&30.34&17.55&16.62&99.23 \\
                                          &                         &7&128&291.57&248.70&53.60&28.93&33.37&32.71&99.82 \\
                                          &                         &8&256&287.74&254.06&53.75&28.55&66.48&66.18&97.39 \\ \cmidrule{3-11}

     \multirow{6}{*}{Face-Iris}           &                         &3&8&494.24&272.27&76.04&49.03&3.56&2.41&100.00 \\ 
                                          &                         &4&16&456.16&279.74&73.01&45.25&6.74&4.84&100.00 \\
                                          &                         &5&32&424.87&277.07&72.77&42.15&12.57&9.42&100.00 \\
                                          &                         &6&64&426.95&285.31&70.66&42.36&25.83 & 19.16&99.98 \\
                                          &                         &7&128&414.45&286.52& 69.54&41.12& 50.55 & 38.02&99.45 \\
                                          &                         &8&256&422.75&280.94&69.81&41.94&103.74&74.51&92.62 \\ \cmidrule{3-11}

     \multirow{6}{*}{Face-Fingerprint-Iris}&                        &3&8&454.35&261.03&70.97&45.07&3.13&2.29&100.00 \\
                                           &                         &4&16&400.79&266.70& 66.22&39.76&5.72&4.55& 100.00 \\
                                           &                        &5&32&372.04&262.79& 62.98&36.91&10.60&8.94&100.00 \\
                                           &                         &6&64&356.37&266.17& 61.76&35.35& 20.61&17.85&99.11 \\
                                           &                        &7&128&347.06&271.57& 61.37&34.43&40.80&35.86&99.97 \\
                                           &                         &8&256&345.46&271.78&61.23&34.27&81.94&71.37&99.24 \\ \cmidrule{1-11}

  \multirow{6}{*}{Face-Fingerprint}&\multirow{24}{*}{BioHashing}    &3&8 &454.07&270.89&71.92&45.05&3.43&2.28&100.00\\ 
                                          &                         &4&16&400.53&267.95& 66.32&39.74& 5.92&4.48&100.00  \\
                                          &                         &5&32&346.58&264.32&60.60&34.38&10.27&8.55&100.00 \\
                                          &                         &6&64&320.48&260.45&57.63 & 31.79&18.44&16.27&100.00 \\
                                          &                         &7&128&297.68&255.27&54.86&29.53& 32.83&31.40& 98.89 \\
                                          &                         &8&256&271.14 & 245.79& 51.28 & 26.907& 57.73& 58.92&98.37   \\ \cmidrule{3-11}
                                          
\multirow{6}{*}{Iris-Fingerprint}          &                        &3&8& 428.38& 263.15&68.60 & 42.50&3.24& 2.21& 100.00   \\
                                           &                        &4&16& 363.00 & 258.59&61.67 & 36.01& 5.27 & 4.25&100.00   \\
                                           &                        &5&32& 312.50 & 251.76 &55.98 & 31.00&9.09&7.99& 100.00 \\
                                           &                        &6&64&278.53 & 246.08&  52.04 & 27.63& 15.74 & 15.15& 100.00 \\
                                           &                        &7&128&271.60& 244.73& 51.22 & 26.94&29.72 & 29.62& 98.95    \\
                                           &                        &8&256&249.93 & 237.19& 48.33 & 24.79&52.22 & 55.93&98.01  \\ \cmidrule{3-11}

\multirow{6}{*}{Face-Iris}                 &                        &3&8&520.65 & 288.21&80.24 & 51.65&3.96 & 2.46&100.00    \\
                                           &                        &4&16& 468.75 & 286.00&74.88& 46.50&7.19&4.84&100.00  \\
                                           &                        &5&32&450.12 & 290.18&73.44& 44.65&14.01 & 9.69&100.00  \\
                                           &                        &6&64&433.83& 285.35&71.35 & 43.04&27.07&18.99&100.00    \\
                                           &                        &7&128&432.10 & 289.35& 71.57& 42.87& 53.87 & 38.11& 98.46     \\
                                           &                        &8&256&430.76 & 283.90&70.90 & 42.73&106.64 & 74.41&92.51  \\ \cmidrule{3-11}

\multirow{6}{*}{Face-Fingerprint-Iris}     &                        &3&8&465.20& 275.64&73.50 & 46.15&3.61 & 2.31&100.00     \\
                                           &                        &4&16& 409.05 & 275.37&67.90 & 40.58&6.15 & 4.57&100.00   \\
                                           &                        &5&32& 374.94 & 273.39&64.32 & 37.20&11.30 & 8.87&100.00   \\
                                           &                        &6&64& 349.57 & 270.70& 61.53 & 34.68&20.74 & 17.20&100.00     \\
                                           &                        &7&128 & 329.18 & 267.62& 59.21 & 32.66& 38.32 & 33.67&98.99  \\
                                           &                        &8&256& 311.71 & 263.34& 57.05 & 30.92&70.88 & 65.28&98.53      \\ \cmidrule{1-11}

 \multirow{6}{*}{Face-Fingerprint}        &\multirow{24}{*}{IoM-GRP}&3&8& 526.26 & 254.99&77.51 & 52.21&2.82 & 2.22&100.00   \\ 
                                          &                         &4&16& 465.68 & 255.13&71.51&46.20&4.46 & 4.05&100.00    \\
                                          &                         &5&32& 428.41 & 253.00&67.60&42.50&7.69 & 7.79&100.00 \\
                                          &                         &6&64& 377.47 & 244.82& 61.74 & 37.45&13.21 & 14.72&100.00 \\
                                          &                         &7&128& 340.09 & 241.42&57.69 & 33.74&26.12 & 28.99&100.00\\
                                          &                         &8&256& 301.70 & 245.04&54.24 & 29.93&52.38 & 58.01&100.00 \\ \cmidrule{3-11}

\multirow{6}{*}{Iris-Fingerprint}         &                         &3&8& 502.29 & 251.35&74.76 & 49.83&2.584 & 2.12&100.00 \\
                                          &                         &4&16& 441.14 & 247.62 &68.33 & 43.76& 4.03 & 3.85&100.00    \\
                                          &                         &5&32&  392.84 & 244.02& 63.18 & 38.97& 6.77 & 7.19 &100.00  \\
                                          &                         &6&64&  356.30 & 241.50&59.31 & 35.35& 12.43 & 14.20&100.00 \\
                                          &                         &7&128&  318.01 & 235.93&54.95 & 31.55&24.23 & 27.70&100.00 \\
                                          &                         &8&256&  272.87 & 232.65 &  50.15 & 27.07&46.81 & 54.03&100.00 \\ \cmidrule{3-11}

\multirow{6}{*}{Face-Iris}                &                         &3&8&515.90 & 276.97&78.66 & 51.18&3.58 & 2.48&100.00     \\
                                          &                         &4&16& 459.13 & 278.06& 73.13 & 45.54&6.51 & 4.87&100.00  \\
                                          &                         &5&32& 443.98 & 280.69&71.89 & 44.05& 12.92 & 9.72&100.00  \\
                                          &                         &6&64& 422.37 & 286.35&70.31 & 41.90 &25.44 & 19.31&100.00   \\
                                          &                         &7&128& 418.49 & 285.40& 70.72 & 41.52&51.30 & 38.08&100.00    \\
                                          &                         &8&256&424.41 & 288.45&69.83 & 42.10& 105.16 & 76.07&99.99   \\ \cmidrule{3-11}

\multirow{6}{*}{Face-Fingerprint-Iris}    &                         &3&8&520.76 & 258.87&77.35 & 51.66&2.94 & 2.29& 100.00    \\
                                          &                         &4&16&461.46 & 258.30&71.40 & 45.78& 4.90 & 4.35&100.00     \\
                                          &                         &5&32&  426.69 & 256.38&67.77 & 42.33&8.93 & 8.45&100.00  \\
                                          &                         &6&64& 390.98 & 256.74&64.26 & 38.79&16.38 & 16.52&100.00   \\
                                          &                         &7&128&  365.32 & 256.47&61.69 & 36.24&32.95 & 32.94&100.00    \\
                                          &                         &8&256& 336.67 & 259.74 &59.17 & 33.40&67.22 & 65.12&100.00   \\ 
              
    \bottomrule
	\end{tabular}
		\begin{tablenotes}
  \item $k$: Length of the frequent binary pattern, \#Comb: Number of possible combinations to be generated given a $k$, \#Comp: Average number of comparisons, Std\_comp: Standard deviation across the number of comparisons carried out per subject, W$_l$: Lower bound of the computational workload reduction estimated on the average number of comparisons computed per subject, W$_u$: Upper bound of the computational workload reduction, \#Visited-patterns: Average number of binary patterns visited from the probe, Std\_bins\_v: Standard deviation across the bins visited per subject, H-R: Hit-Rate.
  \end{tablenotes}
	\end{threeparttable}
	\end{adjustbox}

\end{adjustbox}
\end{table*}

\begin{table*}[!t]
\centering
	\caption{Closed scenario over multi-biometric characteristics. Results are shown for the most ranked-code indexing.}
 \begin{adjustbox}{width=0.8\linewidth}
     	\label{tab:SM_ranked-code-closed}
    \begin{adjustbox}{max width=\linewidth}
    \begin{threeparttable}
    \begin{tabular}{ccccccccccc}
    \toprule
        \textbf{Combination}& \textbf{Approach} & \textbf{k} & \textbf{\#Comb} & \textbf{\#Comp} & \textbf{Std\_comp} & \textbf{W$_u$(\%)} & \textbf{W$_l$(\%)} & \textbf{\#Visited-patterns} &\textbf{Std\_bins\_v}& \textbf{H-R} \\ \cmidrule{1-11}
         \multirow{6}{*}{Face-Fingerprint}&\multirow{24}{*}{Baseline}&3&8&468.57 & 264.51 &72.73&46.48&2.96 & 2.03&100.00\\
                                          &                         &4&16&394.93 & 260.38& 65.01&39.18&5.53 & 4.12 &100.00  \\
                                          &                         &5&32&355.85 & 257.08 &60.80 & 35.30 &10.15 & 8.15 & 99.99  \\
                                          &                         &6&64& 340.12 & 260.56 & 59.59 &  33.74 &19.65 & 16.55 &  99.47  \\
                                          &                         &7&128&332.79 & 254.12 &58.23 & 33.01 & 38.80 & 33.27 & 89.56\\
                                          &                         &8&256& 231.27 & 130.26 &  35.87 &  22.94 & 54.35 & 35.08& 54.95\\ \cmidrule{3-11}

        \multirow{6}{*}{Iris-Fingerprint}&                          &3& 8& 443.98 & 260.32 &69.87 & 44.05 & 2.85 & 1.99 & 100.00   \\
                                          &                         &4&16 & 364.01 & 253.13 & 61.22 &  36.11 &5.15 &3.99 & 100.00 \\
                                          &                         &5&32& 330.72 & 251.67 & 57.78 & 32.81 & 9.49 & 7.92 & 100.00     \\
                                          &                         &6&64& 307.02 & 247.93 & 55.05 & 30.46 & 17.73 & 15.54 & 99.72   \\
                                          &                         &7&128& 307.88 & 250.00 & 55.35 & 30.54 & 35.55 & 32.35 & 91.57 \\
                                          &                         &8&256& 216.20 & 133.31 & 34.67 & 21.45 & 50.76 & 35.53 &  59.49   \\ \cmidrule{3-11}

     \multirow{6}{*}{Face-Iris}&                                    &3&8&  512.58 & 276.62 & 78.29 &  50.85 &  3.46 & 2.19 & 100.00    \\
                                          &                         &4&16 & 454.31  & 277.26 & 72.58 &  45.07 & 6.70 & 4.44 & 100.00 \\
                                          &                         &5&32& 433.38 & 283.41& 71.11 &  42.99 &  13.03 & 9.15 &  99.24  \\
                                          &                         &6&64& 422.60 & 285.52 & 70.25 & 41.92 &  25.450 & 18.33 & 99.97   \\
                                          &                         &7&128& 410.60 & 265.99 & 67.12 & 40.73 &  49.55 & 35.18 & 86.05  \\
                                          &                         &8&256& 261.73 & 115.78 & 37.45 &  25.96 & 64.17  & 31.81 & 43.23     \\ \cmidrule{3-11}

     \multirow{6}{*}{Face-Fingerprint-Iris}&                        &3&8& 474.58 & 267.71 & 73.64 &  47.08 &  3.13 & 2.09 & 100.00   \\
                                          &                         &4&16 &  403.53 & 267.89 &66.61 & 40.03 & 5.86 & 4.26 & 100.00  \\
                                          &                         &5&32& 368.78 & 265.02 & 62.88 &36.59 & 10.75 & 8.38 &  99.08   \\
                                          &                         &6&64& 365.12 & 267.47 & 62.76 &36.22 &  21.11 & 16.78 & 99.96  \\
                                          &                         &7&128& 351.96 & 251.22 & 59.84 &34.92 & 40.97 & 32.94 & 83.39    \\
                                          &                         &8&256& 166.22 & 74.26 & 23.86 & 16.49 &  39.68 & 19.39 & 34.42    \\ \cmidrule{1-11}

   \multirow{6}{*}{Face-Fingerprint}&\multirow{24}{*}{BioHashing}   &3&8& 493.10 & 260.53 & 74.77 &48.92 & 2.90 & 1.96 & 100.00   \\
                                          &                         &4&16& 383.74 & 258.33 & 63.70  & 38.07 & 5.36 & 3.95 & 100.00 \\
                                          &                         &5&32& 331.37 & 246.72 & 57.35 & 32.87 &  9.46 & 7.53 & 100.00\\
                                          &                         &6&64&  312.80 & 247.64 &  55.60 & 31.03 & 17.31 & 14.80 & 99.72  \\
                                          &                         &7&128& 313.66 & 256.36 & 56.55 & 31.12 & 33.38 & 31.21 &  91.70\\
                                          &                         &8&256&  237.60 & 152.52 & 38.70 & 23.57 &  48.98 & 36.01 &  61.50   \\ \cmidrule{3-11}

        \multirow{6}{*}{Iris-Fingerprint}&                          &3& 8& 476.88 & 255.72 & 72.68 & 47.31 & 2.76 & 1.91 & 100.00   \\
                                          &                         &4&16 & 363.95 & 255.67 & 61.47 & 36.11 &  5.02 & 3.88 & 100.00 \\
                                          &                         &5 & 32 & 311.60 & 243.14 & 55.03 & 30.91 & 8.78 & 7.36 & 100.00    \\
                                          &                         &6&64& 287.23 & 236.06 & 51.91 & 28.50 & 15.73  & 13.98 &  99.73      \\
                                          &                         &7&128& 286.75 & 249.71 & 53.22 & 28.45 & 30.06 & 29.86 & 93.65  \\
                                          &                         &8&256& 226.82 & 155.96 & 37.97 & 22.50 & 46.20 & 36.43 & 64.62  \\ \cmidrule{3-11}

     \multirow{6}{*}{Face-Iris}&                                    &3&8& 531.98 & 281.86 &  80.74 & 52.78 &3.64 & 2.28 & 100.00      \\
                                          &                         &4&16 & 478.46 & 280.50 & 75.29 & 47.47 &  7.18 & 4.52 & 100.00 \\
                                          &                         &5&32& 449.19 & 283.99 & 72.74 &  44.56 & 13.76 & 9.11 & 100.00 \\
                                          &                         &6&64& 437.91 & 285.88 & 71.80 & 43.44 & 26.83 & 18.38 & 99.28      \\
                                          &                         &7&128& 424.86 & 267.69 & 68.71 &  42.15 & 51.78 & 35.17 & 86.32 \\
                                          &                         &8&256& 267.15 & 115.05 & 37.92 & 26.50 & 65.82 & 31.76 &  42.74    \\ \cmidrule{3-11}

     \multirow{6}{*}{Face-Fingerprint-Iris}&                        &3&8&499.24 & 268.09 & 76.12 &49.53 & 3.16 & 2.10 & 100.00 \\
                                          &                         &4&16 & 408.68 & 268.47 & 67.18 & 40.54 &5.84 & 4.18 & 100.00 \\
                                          &                         &5&32& 367.51 & 259.97 &62.25 & 36.46 & 10.67 & 8.02 & 100.00  \\
                                          &                         &6&64& 349.82 & 262.88 &60.78 & 34.70 & 19.80 & 15.99 & 99.43  \\
                                          &                         &7&128& 337.37 & 256.65 &  58.93 &33.47 & 38.00 & 32.00 & 86.71  \\
                                          &                         &8&256&  179.98 & 82.90 & 26.08 & 17.85 & 39.28 & 20.33 & 37.37     \\ \cmidrule{1-11}

   \multirow{6}{*}{Face-Fingerprint}&\multirow{24}{*}{IoM-GRP}      &3&8& 499.07 & 260.94 & 75.40 & 49.51 &  2.80 & 1.94 & 100.00       \\
                                          &                         &4&16& 439.98 &  260.40 & 69.48 &43.65 &4.75 & 3.81 & 100.00  \\
                                          &                         &5&32& 389.56 & 254.10 &  63.85 & 38.65 &8.43 & 7.27& 100.00 \\
                                          &                         &6&64& 354.93 & 247.04 & 59.72 & 35.21 &15.64 & 13.98 & 100.00      \\
                                          &                         &7&128& 331.16 & 241.98 & 56.86 &32.85 & 30.14 & 27.23 & 99.41   \\
                                          &                         &8&256& 303.24 & 239.92 & 53.88& 30.08 & 56.64 & 54.00 & 99.98    \\ \cmidrule{3-11}

        \multirow{6}{*}{Iris-Fingerprint}&                          &3& 8& 476.04 & 255.34 & 72.56 & 47.23 &2.58 & 1.87 & 100.00     \\
                                          &                         &4&16 & 406.76 & 253.44 & 65.50 & 40.35 & 4.44 & 3.72 & 100.00  \\
                                          &                         &5 & 32 & 355.77 & 241.73 &59.28 & 35.29 &  7.63 & 6.90 & 100.00   \\
                                          &                         &6&64& 318.98 & 238.15 & 55.27 & 31.64 & 14.05 & 13.35 & 100.00      \\
                                          &                         &7&128& 294.95 & 235.50 & 52.62 & 29.26 & 26.80 & 26.42 & 100.00   \\
                                          &                         &8&256& 275.50 & 233.84 & 50.53 &  27.33 & 51.51 & 52.19 &99.64  \\ \cmidrule{3-11}

     \multirow{6}{*}{Face-Iris}&                                    &3&8&  515.87 & 278.99 &  78.86 &  51.18 & 3.48 & 2.22 & 100.00     \\
                                          &                         &4&16 & 454.99 & 281.78 & 73.09 &  45.14 &  6.68 & 4.60 & 100.00\\
                                          &                         &5&32& 439.28 & 283.04 & 71.66 & 43.58 &13.25 & 9.19 & 100.00  \\
                                          &                         &6&64& 423.42 & 282.27 & 70.01 & 42.01 & 26.06 & 18.13 & 100.00   \\
                                          &                         &7&128& 422.17 & 284.60 & 70.12 & 41.88 & 51.90 & 36.58 & 100.00  \\
                                          &                         &8&256& 430.86 & 287.75 & 71.29 & 42.74 & 105.33 & 73.94 & 99.10    \\ \cmidrule{3-11}

     \multirow{6}{*}{Face-Fingerprint-Iris}&                        &3&8& 498.55 & 266.69 & 75.92 & 49.46 & 3.12 & 2.16 &100.00 \\
                                          &                         &4&16 &  433.85 & 262.62 & 69.09 & 43.04 & 5.31 & 4.14 & 100.00  \\
                                          &                         &5&32& 387.48 & 263.67 & 64.60 & 38.44 & 9.59 & 8.08 & 100.00    \\
                                          &                         &6&64&361.49 & 260.48 &  61.70 & 35.86 & 18.39 & 15.76 & 100.00 \\
                                          &                         &7&128& 344.41 & 255.47 & 59.51 & 34.17 & 35.31 & 30.50 & 99.98  \\
                                          &                         &8&256&337.05 & 258.33 & 59.07 & 33.44 & 69.59 & 61.75 &99.98 \\   

    \bottomrule
	\end{tabular}
		\begin{tablenotes}
  \item $k$: Length of the frequent binary pattern, \#Comb: Number of possible combinations to be generated given a $k$, \#Comp: Average number of comparisons, Std\_comp: Standard deviation across the number of comparisons carried out per subject, W$_l$: Lower bound of the computational workload reduction estimated on the average number of comparisons computed per subject, W$_u$: Upper bound of the computational workload reduction, \#Visited-patterns: Average number of binary patterns visited from the probe, Std\_bins\_v: Standard deviation across the bins visited per subject, H-R: Hit-Rate.
  \end{tablenotes}
	\end{threeparttable}
	\end{adjustbox}

\end{adjustbox}
\end{table*}

\begin{table*}[!t]
\centering
	\caption{Closed scenario over multi-biometric characteristics. Results are shown for the XOR-code indexing.}
 \begin{adjustbox}{width=0.8\linewidth}
     	\label{tab:SM_closed_set_xor_codes}
    \begin{adjustbox}{max width=\linewidth}
    \begin{threeparttable}
    \begin{tabular}{ccccccccccc}
    \toprule
        \textbf{Combination}& \textbf{Approach} & \textbf{k} & \textbf{\#Comb} & \textbf{\#Comp} & \textbf{Std\_comp} & \textbf{W$_u$(\%)} & \textbf{W$_l$(\%)} & \textbf{\#Visited-patterns} &\textbf{Std\_bins\_v}& \textbf{H-R} \\ \cmidrule{1-11}
         \multirow{6}{*}{Face-Fingerprint}&\multirow{24}{*}{Baseline}&3&8& 520.07 & 293.84 &80.75&51.59 & 4.05 & 2.40 & 100.00   \\
                                          &                         &4&16& 517.54 & 293.18 & 80.43 & 51.34 & 8.17 & 4.72 & 100.00 \\
                                          &                         &5&32& 499.80 & 297.56 & 79.10 & 49.58 &15.82 & 9.54 & 100.00   \\
                                          &                         &6&64& 492.15 & 295.93 & 78.18  & 48.82 & 31.14 & 18.87 & 100.00 \\
                                          &                         &7&128& 496.76 & 296.40 & 78.69 & 49.28 & 62.87 & 37.71 & 100.00  \\
                                          &                         &8&256& 498.75 & 293.08 & 78.55 & 49.48 & 125.95 & 74.72 & 100.00    \\ \cmidrule{3-11}
        \multirow{6}{*}{Iris-Fingerprint} &                         &3&8& 517.01 & 295.39 & 80.60 & 51.29 & 4.04 & 2.40 & 100.00 \\
                                          &                         &4&16& 498.61 & 295.85 &80.60 & 51.29 & 4.04 & 2.40 & 100.00    \\
                                          &                         &5&32& 498.61 & 295.85 & 78.82 & 49.46 & 7.87  & 4.75 & 100.00    \\
                                          &                         &6&64& 491.17 & 296.09 & 78.10 & 48.72 & 31.08 & 18.87 &100.00\\
                                          &                         &7&128& 493.87 & 293.75 & 78.14 & 49.00 & 62.49 & 37.37 & 100.00    \\
                                          &                         &8&256& 496.22 & 291.67 & 78.20 & 78.16 & 125.35 & 74.29 & 100.00   \\ \cmidrule{3-11}
        \multirow{6}{*}{Face-Iris}        &                         &3&8& 549.39 & 293.53 & 83.62 & 54.50 & 4.28 & 2.41 & 100.00 \\
                                          &                         &4&16& 521.28 & 292.55 & 80.74 & 51.71 & 8.23 & 4.69 & 100.00   \\
                                          &                         &5&32& 522.28 & 293.82 &78.09 &51.81 & 16.51 & 9.42 & 100.00    \\
                                          &                         &6&64& 496.82 & 290.28 &  78.99 & 49.29 &31.43 & 18.51 & 100.00 \\
                                          &                         &7&128& 504.61 & 291.94 & 79.02 & 50.06 & 63.87 & 37.16 & 100.00   \\
                                          &                         &8&256& 507.76 & 291.47 &  79.29 &  50.37 & 128.42 & 74.13 & 100.00  \\ \cmidrule{3-11}
     
\multirow{6}{*}{Face-Fingerprint-Iris}    &                         &3&8&  598.49 & 255.57 &84.73 & 59.37 & 3.28 & 2.26 & 100.00  \\
                                          &                         &4&16& 562.70 & 259.71 &81.59 & 55.82 & 5.67 & 4.26 & 100.00    \\
                                          &                         &5&32&  544.51 & 267.74 & 80.58 & 54.02 &10.21 & 8.35 & 100.00    \\
                                          &                         &6&64& 531.58 & 273.48 & 79.87 & 52.74 & 18.78 & 16.04 & 100.00 \\
                                          &                         &7&128&  511.69 & 281.83 & 78.72 &50.76 & 34.93 & 31.25 & 100.00     \\
                                          &                         &8&256& 510.33 & 285.05 & 78.91 & 50.63 & 66.46 & 59.59 & 100.00   \\ \cmidrule{1-11}

  \multirow{6}{*}{Face-Fingerprint}&\multirow{24}{*}{BioHashing}    &3&8&  543.74 & 296.98 & 83.40 & 53.94 & 4.32 & 2.37 & 100.00  \\
                                          &                         &4&16& 518.40  & 296.38 & 80.83 & 51.43 &8.18 & 4.74 & 100.00 \\
                                          &                         &5&32& 503.68 & 296.37 & 78.00 &  49.97 &15.96 & 9.44 & 100.00    \\
                                          &                         &6&64& 510.84 & 296.79 & 78.30 &  50.68 &  32.35 & 18.87 & 100.00 \\
                                          &                         &7&128& 494.92 & 295.32 & 78.40 &  49.10 & 62.64 & 37.52 & 100.00 \\
                                          &                         &8&256 &500.17 & 294.16 &78.80 & 49.62 &  126.40 & 74.82 & 100.00 \\ \cmidrule{3-11}

  \multirow{6}{*}{Iris-Fingerprint}       &                         &3&8&519.83 & 296.94 & 81.03 & 51.57 & 4.13 & 2.36 & 100.00  \\
                                          &                         &4&16& 503.37 & 296.88 & 79.39 & 49.94 & 7.94 & 4.77 & 100.00 \\
                                          &                         &5&32& 490.34 & 301.43 &78.55 &48.64 & 15.54 & 9.60 & 100.00  \\
                                          &                         &6&64& 496.44 & 296.14 & 78.63 & 49.25 & 31.46 & 18.82 & 100.00 \\
                                          &                         &7&128& 497.41 & 295.85 &  78.70 & 49.35 &  62.97 & 37.57 & 100.00  \\
                                          &                         &8&256&500.40 & 296.51 & 79.06 & 49.64 & 126.53 & 75.39 & 100.00  \\ \cmidrule{3-11}

  \multirow{6}{*}{Face-Iris}              &                         &3&8& 563.83 & 288.70 & 84.58 & 55.94 & 4.46 & 2.33 & 100.00   \\
                                          &                         &4&16& 537.13 & 288.53 &81.91 & 53.29 & 8.50 & 4.62 & 100.00    \\
                                          &                         &5&32& 511.60 & 290.60 &79.58 & 50.75 & 16.21 & 9.29 & 100.00 \\
                                          &                         &6&64& 511.54 & 293.12 & 79.83 & 50.75 & 32.42 & 18.67 & 100.00 \\
                                          &                         &7&128& 496.90 & 291.27 &  78.19 & 49.30 & 62.93 & 37.03 & 100.00  \\
                                          &                         &8&256& 503.36 & 292.00 & 78.91 & 49.94 & 127.35 & 74.36 & 100.00\\ \cmidrule{3-11}

  \multirow{6}{*}{Face-Fingerprint-Iris}  &                         &3&8& 605.01 & 257.52 & 85.57 &  60.02 & 3.38 & 2.15 & 100.00\\
                                          &                         &4&16& 571.38 & 261.35 & 82.61 & 56.68 &  5.53 & 4.06 & 100.00    \\
                                          &                         &5&32& 548.28 & 265.26 & 80.71 & 54.39 & 9.80 & 7.90 & 100.00  \\
                                          &                         &6&64& 529.85 & 275.20 &79.87 &  52.56 &18.42 & 15.40 & 100.00     \\
                                          &                         &7&128& 515.86 & 280.33 & 78.77 &51.18 &33.39 & 29.49 & 100.00  \\
                                          &                         &8&256& 510.10 & 283.95 &  78.99 & 50.61 & 62.91 & 56.82 & 100.00     \\ \cmidrule{1-11}

  \multirow{6}{*}{Face-Fingerprint}  & \multirow{24}{*}{IoM-GRP}    &3&8& 542.50 & 294.06 & 82.99 & 53.82 & 4.04 & 2.49 & 100.00 \\
                                          &                         &4&16&514.80 & 295.57 &80.39 &  51.07 & 7.83 & 4.84 & 100.00    \\
                                          &                         &5&32& 510.59 & 295.16 & 79.94 & 50.65 & 15.74 & 9.57 & 100.00   \\
                                          &                         &6&64& 505.31 & 295.53 & 79.45 & 50.13 &  31.62 & 18.97 & 100.00      \\
                                          &                         &7&128& 506.47 & 293.21 & 79.33 & 50.25 & 63.84 & 37.48 & 100.00  \\
                                          &                         &8&256& 495.76 & 294.41 &78.39 & 49.18 & 125.45 & 74.84 & 100.00    \\ \cmidrule{3-11}

    \multirow{6}{*}{Iris-Fingerprint}     &                         &3&8&534.17& 291.27 & 81.89 & 52.99 & 3.95  & 2.49 & 100.00   \\
                                          &                         &4&16& 500.11 & 301.36 & 79.51 & 49.61 & 7.56 & 4.92 & 100.00   \\
                                          &                         &5&32& 490.35 & 302.47 & 78.20 & 48.65 &15.03 & 9.86 & 100.00     \\
                                          &                         &6&64& 491.63 & 300.09 & 78.54 & 48.77 &30.73 & 19.28 & 100.00       \\
                                          &                         &7&128& 494.63 & 296.25 & 78.46 & 49.07 & 62.31 & 37.77 & 100.00  \\
                                          &                         &8&256& 493.13 & 296.28 & 78.31 & 48.92 & 124.63 & 75.42 & 100.00    \\ \cmidrule{3-11}

    \multirow{6}{*}{Face-Iris}     &                                &3&8&  562.71 & 286.81 &84.28 & 55.82 & 4.32 & 2.40 & 100.00  \\
                                          &                         &4&16&  516.24 & 289.89 & 79.97 &51.21 & 8.11 & 4.68 & 100.00  \\
                                          &                         &5&32& 512.60 & 293.88 & 80.01 & 50.85 & 16.22 & 9.41 & 100.00  \\
                                          &                         &6&64&  502.50 & 292.35 &78.85 & 49.85 & 31.83 & 18.63 & 100.00       \\
                                          &                         &7&128&  509.21 & 293.00 & 79.58 & 50.52 &  64.59 & 37.27 & 100.00\\
                                          &                         &8&256& 510.55 & 291.86 &79.60 &  50.65 & 129.55 & 74.15 & 100.00    \\ \cmidrule{3-11}

\multirow{6}{*}{Face-Fingerprint-Iris}    &                         &3&8&  632.55 & 238.54 & 86.42 & 62.75 & 3.33 & 2.35 & 100.00   \\
                                          &                         &4&16& 587.00 & 247.59 & 82.80 & 58.23 &  5.65 & 4.51 & 100.00  \\
                                          &                         &5&32&  554.64 & 255.26 & 80.35 & 55.02 & 10.04 & 8.58 & 100.00\\
                                          &                         &6&64& 534.92 & 266.33 & 79.49 & 53.07 & 17.89 & 16.43 & 100.00       \\
                                          &                         &7&128& 522.69 & 274.77 & 79.11 &  51.85 & 32.78 & 30.98 & 100.00 \\
                                          &                         &8&256& 512.57 & 282.02 & 78.83 & 50.85 & 62.06 & 59.03 & 100.00  \\

    \bottomrule
	\end{tabular}
		\begin{tablenotes}
  \item $k$: Length of the frequent binary pattern, \#Comb: Number of possible combinations to be generated given a $k$, \#Comp: Average number of comparisons, Std\_comp: Standard deviation across the number of comparisons carried out per subject, W$_l$: Lower bound of the computational workload reduction estimated on the average number of comparisons computed per subject, W$_u$: Upper bound of the computational workload reduction, \#Visited-patterns: Average number of binary patterns visited from the probe, Std\_bins\_v: Standard deviation across the bins visited per subject, H-R: Hit-Rate.
  \end{tablenotes}
	\end{threeparttable}
	\end{adjustbox}

\end{adjustbox}
\end{table*}

\begin{table*}[!t]
\centering
	\caption{Open-set results over Baseline across different indexing-schemes.}
 \begin{adjustbox}{width=0.8\linewidth}
     	\label{tab:SM_baseline-open-set-multi-bio}
    \begin{adjustbox}{max width=\linewidth}
    \begin{tabular}{cccccccc}
    \toprule
        \textbf{Method}&\textbf{Biometric-characteristic} & \textbf{Approach} &\textbf{k} &\textbf{\#Bins}& \textbf{W(in\%)}& \textbf{FPIR=0.01(in\%)} & \textbf{FPIR=0.1(in\%)}\\ \cmidrule{1-8}

                 \multirow{3}{*}{Individual}&Face(indexing) & \multirow{3}{*}{Baseline} & 5 & 23 &69.99 &34.90 &34.00 \\ 
                                        
                                            &Fingerprint(indexing) &                    &6&26&43.30  &28.76 &24.86 \\ 

                                               
                                                                                                                                         
                                            &Iris(indexing)  &                          &5&22&67.06&70.59 &49.70 \\ \cmidrule{1-8}

                                        
            
             \multirow{12}{*}{Feature-concatenation}&Face-Fingerprint&\multirow{12}{*}{Baseline}&6&37&58.50 &25.04 &20.87 \\
                                                  &Fingerprint-Face&                         &6&37&58.55 &23.21&20.90 \\
                                                  &Iris-Fingerprint&                         &7&66&52.85 &27.56 &24.03 \\ 
                                                  &Fingerprint-Iris&                         &7&66&52.91&24.15 &22.32 \\
                                                  &Face-Iris&                                &6&45&68.96&27.33 &25.13\\
                                                  &Iris-Face&                                &6&45&68.89&24.75 &23.58\\ 
                                                  &Face-Fingerprint-Iris&                    &7&76&60.16&20.65 &20.14 \\
                                                  &Face-Iris-Fingerprint&                    &7&76&60.25&20.94 &20.19 \\  
                                                  &Fingerprint-Face-Iris&                    &7&76&60.24&20.83 &20.00 \\
                                                  &Fingerprint-Iris-Face&                    &7&76&60.12&20.38 &20.36 \\
                                                  &Iris-Face-Fingerprint&                    &7&76&60.26&20.63 &19.63 \\
                                                  &Iris-Fingerprint-Face&                    &7&76&60.07&21.93 &20.26 \\ \cmidrule{1-8}

           \multirow{12}{*}{Ranked-codes}&        Face-Fingerprint&\multirow{12}{*}{Baseline}&6&36&59.83&21.18&19.85\\
                                                  &Fingerprint-Face&                         &6&36&59.83&21.18&19.85\\
                                                  &Iris-Fingerprint&                         &6&33&55.11&24.59&23.68\\ 
                                                  &Fingerprint-Iris&                         &6&33&55.11&24.59&23.68\\
                                                  &Face-Iris&                                &6&44&70.38&25.15&22.23\\
                                                  &Iris-Face&                                &6&44&70.38&25.15&22.23 \\ 
                                                  &Face-Fingerprint-Iris&                    &6&38&62.97&19.07&18.62\\
                                                  &Face-Iris-Fingerprint&                    &6&38&62.97&19.07&18.62 \\  
                                                  &Fingerprint-Face-Iris&                    &6&38&62.97&19.07&18.62\\
                                                  &Fingerprint-Iris-Face&                    &6&38&62.97&19.07&18.62\\
                                                  &Iris-Face-Fingerprint&                    &6&38&62.97&19.07&18.62\\
                                                  &Iris-Fingerprint-Face&                    &6&38&62.97&19.07&18.62 \\ \cmidrule{1-8}

        \multirow{12}{*}{XOR-codes}&              Face-Fingerprint&\multirow{12}{*}{Baseline} &6&50&77.95&25.84&24.25\\
                                                  &Fingerprint-Face&                          &6&50&77.95&25.84&24.25\\
                                                  &Iris-Fingerprint&                          &6&50&78.05&28.46&27.25\\ 
                                                  &Fingerprint-Iris&                          &6&50&78.05&28.46&27.25\\
                                                  &Face-Iris&                                 &5&25&77.80&26.49&24.63\\
                                                  &Iris-Face&                                 &5&25&77.80&26.49&24.63\\ 
                                                  &Face-Fingerprint-Iris&                     &7&66&81.94&19.38&18.54\\
                                                  &Face-Iris-Fingerprint&                     &7&66&81.94&19.38&18.54\\  
                                                  &Fingerprint-Face-Iris&                     &7&66&81.94&19.38&18.54\\
                                                  &Fingerprint-Iris-Face&                     &7&66&81.94&19.38&18.54\\
                                                  &Iris-Face-Fingerprint&                     &7&66&81.94&19.38&18.54\\
                                                  &Iris-Fingerprint-Face&                     &7&66&81.94&19.38&18.54\\

    \bottomrule
     \end{tabular}
     \end{adjustbox}

\end{adjustbox}
\end{table*}

\begin{table*}[!t]
\centering
	\caption{Open-set results over IoM-GRP across different indexing-schemes.}
 \begin{adjustbox}{width=0.8\linewidth}
     	\label{tab:SM_grp-open-set-multi-bio}

\begin{adjustbox}{max width=\linewidth}
    \begin{tabular}{cccccccc}
    \toprule
        \textbf{Method}&\textbf{Biometric-characteristic} & \textbf{Approach} &\textbf{k} &\textbf{\#Bins}& \textbf{W(in\%)}& \textbf{FPIR=0.01(in\%)} & \textbf{FPIR=0.1(in\%)}\\ \cmidrule{1-8}

                 \multirow{3}{*}{Individual}&Face(indexing) & \multirow{3}{*}{IoM-GRP} & 7 & 93 &71.22 &37.61  &32.10 \\ 
                                        
                                            &Fingerprint(indexing) &                    &7&22&35.86&22.47&19.85 \\

                                            &Iris(indexing)  &                          &7&86&66.63&53.77&47.03 \\ \cmidrule{1-8}

             \multirow{12}{*}{Feature-concatenation}&Face-Fingerprint&\multirow{12}{*}{IoM-GRP}    &8&110&55.30&18.27&17.22 \\
                                                  &Fingerprint-Face&                               &8&110&55.19&18.44&17.98\\
                                                  &Iris-Fingerprint&                               &8&101&51.70&25.76&21.34\\ 
                                                  &Fingerprint-Iris&                               &8&101&51.49&27.14&20.84\\
                                                  &Face-Iris&                                      &8&181&69.81&26.56&24.09\\
                                                  &Iris-Face&                                      &8&181&69.83&26.23&23.83\\ 
                                                  &Face-Fingerprint-Iris&                          &8&132&59.51&18.09&18.01\\
                                                  &Face-Iris-Fingerprint&                          &8&132&59.35&18.82&18.42\\  
                                                  &Fingerprint-Face-Iris&                          &8&132&59.36&17.92&17.61\\
                                                  &Fingerprint-Iris-Face&                          &8&132&59.39&18.38&17.56\\
                                                  &Iris-Face-Fingerprint&                          &8&132&59.56&18.77&17.93\\
                                                  &Iris-Fingerprint-Face&                          &8&132&59.21&18.37&18.01\\ \cmidrule{1-8}

           \multirow{12}{*}{Ranked-codes}&        Face-Fingerprint&\multirow{12}{*}{IoM-GRP} &8&111&55.47&19.27&18.18\\
                                                  &Fingerprint-Face&                         &8&111&55.47&19.27&18.18\\
                                                  &Iris-Fingerprint&                         &8&104&52.23&24.00&20.47\\ 
                                                  &Fingerprint-Iris&                         &8&104&52.23&24.00&20.47\\
                                                  &Face-Iris&                                &7&89&70.40&27.54&23.70\\
                                                  &Iris-Face&                                &7&89&70.40&27.54&23.70 \\ 
                                                  &Face-Fingerprint-Iris&                    &8&131&60.44&17.68&17.33\\
                                                  &Face-Iris-Fingerprint&                    &8&131&60.44&17.68&17.33 \\  
                                                  &Fingerprint-Face-Iris&                    &8&131&60.44&17.68&17.33\\
                                                  &Fingerprint-Iris-Face&                    &8&131&60.44&17.68&17.33\\
                                                  &Iris-Face-Fingerprint&                    &8&131&60.44&17.68&17.33\\
                                                  &Iris-Fingerprint-Face&                    &8&131&60.44&17.68&17.33 \\ \cmidrule{1-8}

        \multirow{12}{*}{XOR-codes}&              Face-Fingerprint&\multirow{12}{*}{IoM-GRP}  &8&200&78.21&22.34&21.98\\
                                                  &Fingerprint-Face&                          &8&200&78.21&22.34&21.98\\
                                                  &Iris-Fingerprint&                          &5&25&76.93&25.95&23.81\\ 
                                                  &Fingerprint-Iris&                          &5&25&76.93&25.95&23.81\\
                                                  &Face-Iris&                                 &6&50&78.02&27.46&26.77\\
                                                  &Iris-Face&                                 &6&50&78.02&27.46&26.77\\ 
                                                  &Face-Fingerprint-Iris&                     &8&121&83.54&17.94&17.46\\
                                                  &Face-Iris-Fingerprint&                     &8&121&83.54&17.94&17.46\\  
                                                  &Fingerprint-Face-Iris&                     &8&121&83.54&17.94&17.46\\
                                                  &Fingerprint-Iris-Face&                     &8&121&83.54&17.94&17.46\\
                                                  &Iris-Face-Fingerprint&                     &8&121&83.54&17.94&17.46\\
                                                  &Iris-Fingerprint-Face&                     &8&121&83.54&17.94&17.46\\

    \bottomrule
     \end{tabular}
     \end{adjustbox}
\end{adjustbox}
\end{table*}

\begin{figure*}[!t]
\centering
\subfloat[Feature-concatenation]{\includegraphics[width=0.25\linewidth]{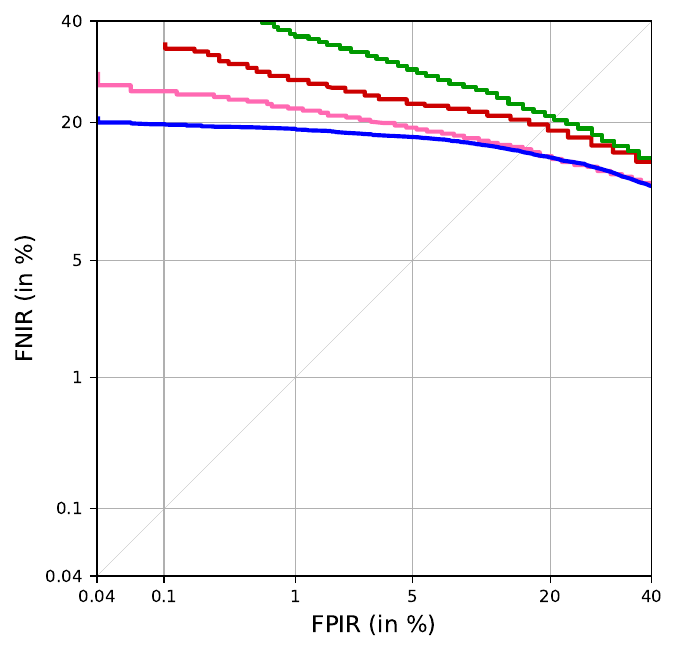}}%
\subfloat[Ranked-codes]{\includegraphics[width=0.25\linewidth]{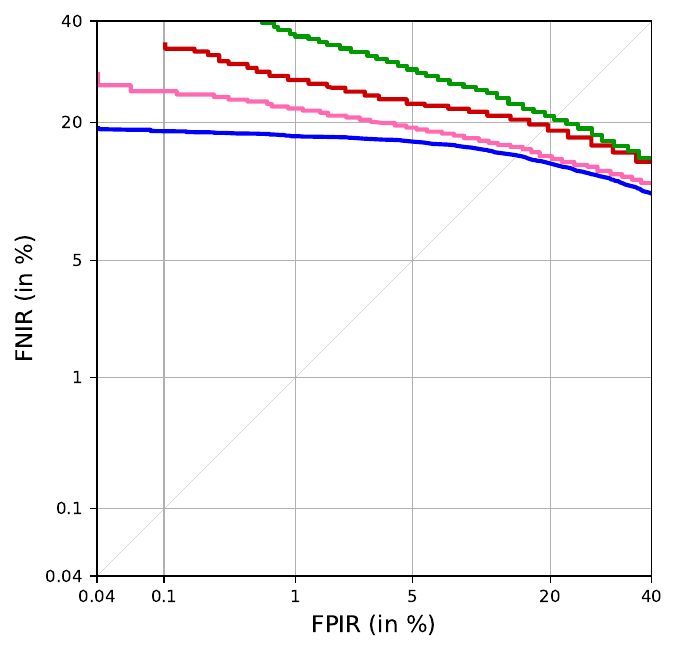}}%
\subfloat[XOR-codes]{\includegraphics[width=0.32\linewidth]{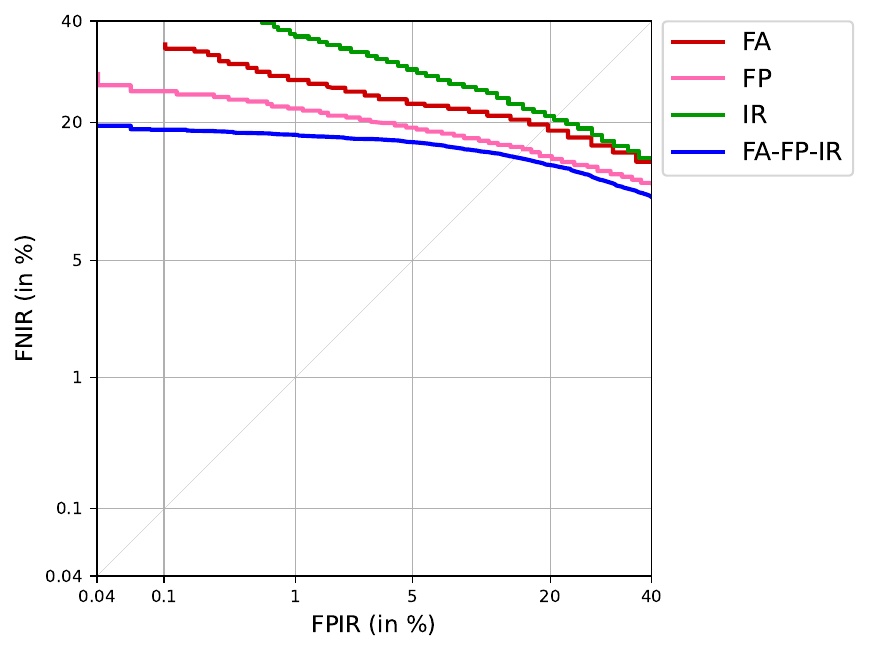}}%
\caption{Best results over open-set scenario are reported on the Baseline for different multi-biometric approaches \wrt their uni-modal approaches.}
\label{fig:SM_open-set-baseline}
\end{figure*}

{\small
\bibliographystyle{IEEEtran}
\bibliography{IEEEabrv}
}

\end{document}